\documentclass{article}

\usepackage{arxiv}

\usepackage[utf8]{inputenc} 
\usepackage[T1]{fontenc}    
\usepackage{hyperref}       

\usepackage[english]{babel}
\usepackage[square,numbers, sort&compress]{natbib}
\usepackage{tabularx}
\usepackage{url}            
\usepackage{booktabs}       
\usepackage{amsfonts}       
\usepackage{amsmath}       
\usepackage{nicefrac}       
\usepackage{microtype}      
\usepackage{lipsum}		
\usepackage{physics}
\usepackage{graphicx}
\usepackage{doi}

\newtheorem{problem}{Problem}

\title{A Short Review on the Maximum Clique Problem Algorithms with Classical,
AI, and Quantum Methods}

\author{	\href{ https://orcid.org/0000-0002-2311-4380}{\includegraphics[scale=0.06]{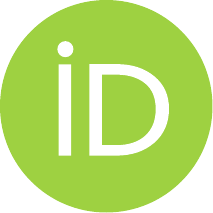}\hspace{1mm} Raffaele Marino}\\
	Dipartimento di Fisica e Astronomia\\
	Università degli studi di Firenze\\
	Via Giovanni Sansone 1, 50019 \\
	Sesto Fiorentino, Florence, Italy \\
	\texttt{raffaele.marino@unifi.it} \\
	\And
         \href{https://orcid.org/0000-0002-8191-3375}{\includegraphics[scale=0.06]{orcid.pdf}\hspace{1mm}Lorenzo Buffoni}\\
	Dipartimento di Fisica e Astronomia\\
	Università degli studi di Firenze\\
	Via Giovanni Sansone 1, 50019 \\
	Sesto Fiorentino, Florence, Italy \\
	\texttt{lorenzo.buffoni@unifi.it} \\
 	\And
         \href{https://orcid.org/0000-0003-3060-0296}{\includegraphics[scale=0.06]{orcid.pdf}\hspace{1mm}Bogdan Zavalnij*}\\
	HUN-REN R\'enyi Institute of Mathematics\\
	H-1053, Re\'altanoda u. 13-15\\
	Budapest, Hungary\\
    *corresponding author\\
	\texttt{bogdan@renyi.hu} \\
}

\renewcommand{\shorttitle}{\textit{arXiv} Template}

\usepackage{xcolor}

\begin{document}
\maketitle

\begin{abstract}

This manuscript provides a comprehensive review of the Maximum Clique Problem, a computational problem that involves finding subsets of vertices in a graph that are all pairwise adjacent to each other. As such, this review is a continuation of the series of previous reviews from 1994, 1999 and 2014.  The manuscript covers in a simple way classical algorithms and includes a review of recent developments in graph neural networks and quantum algorithms. 
\end{abstract}


\section{Introduction} \label{sec::intro}
Imagine you are at a party with $N$ members, and each participant knows some of the members personally. Your goal is to identify the largest group of friends among the $N$ members, meaning the members who know each other. This is known as the Maximum Clique Problem (MCP).

The MCP is a computational problem that involves finding the largest subset of vertices in a graph that are all pairwise adjacent to each other. In the context of our example, it is to discover the largest group of friends at the party who all know each other. This optimization problem is NP-hard.

The term  \textit{clique} originates in social sciences, where complete sub-graphs are used to model social cliques (groups of people who all know each other)\citep{luce1949method}. The first algorithm to solve this sociological problem can be found in Harary et al.\cite{harary1957procedure}. 

Clique-finding algorithms have been used in many scientific disciplines including medicine, biology, mathematics, coding theory, economics, financial markets, wireless networks, telecommunications, cryptography, cybersecurity, social science analysis, and physics, among others\citep{okamoto2020finding,meng2011molecular, das2023improved, strickland2005optimal, Konc2010, Konc2020, ROZMAN2024, makarenkov2023inferring,heal2023p,Corradi1990, lagarias1992keller,brakensiek2022resolution, ashlock2012synthesis, battail2019error, etzion1998greedy,boginski2005statistical,ganesan2023structured, dai2019joint, zhu2021practical, douik2020tutorial, rajsbaum2023distributed, zhong2023designing,lu2023atomic, grandi2022complexity, szabo2022b, ZHONG202320, NASIRIAN2020461, RYSZ2018155}.

The clique problem is related to some important combinatorial optimization problems, such as Graph Coloring, Maximum Independent Set, Minimum Vertex Cover, Graph Clustering, and many others\citep{angelini2023limits, angelini2023stochastic,schaeffer2007graph, guo2022graph, Bota2015, marino2016backtracking, marino2023large, szabo2022a, jovanovic2023fixed, ZHOU202063, PINTO2018849, ZHOU201741}. All of these problems can be directly formulated as an MCP or require finding a maximum clique.

Given its fundamental importance in science and its vast number of applications, several influential surveys on the MCP have been published over the years, with the three most notable ones dating back to 1994, 1999, and 2014\cite{pardalos1994maximum, bomze1999maximum, wu2015review}.

There has been significant advancement in algorithmic research, particularly in the areas of artificial intelligence (AI) and quantum algorithms. Various experiments and numerical analyses have been conducted on the MCP, which necessitates a comprehensive survey. Therefore, this paper aims to fill this gap by providing a detailed review of recent literature on different approaches to the MCP. Our review will cover classical algorithms and, to the best of our knowledge, include AI and quantum algorithms.

This paper is organized into several sections. First, we present the mathematical formulation of the Maximum Clique Problem and the benchmark collections used for testing the algorithms. Second, we describe the classical algorithms for solving the Clique Problem. Third, we provide a general discussion of Graph Neural Networks. Fourth, we review progress by the Quantum Computing community in applying and solving the Maximum Clique Problem and related variants. Finally, we conclude our short review with a discussion and outlines some open questions for these approaches to this problem.

\section{Maximum Clique Problem}\label{sec::MCP}

In this work, if not stated otherwise, we will refer to simple, finite, undirected, unweighted graphs. A graph $G$ is a collection of vertices (also known as nodes or sites) connected by edges (also known as links or arcs). The set of nodes is finite. Formally, a graph is $G(\mathcal{V}, \mathcal{E})$. $\mathcal{V}=\{1,2,\ldots, N\}$ represents the set of vertices and $\mathcal{E}$ represents the set of edges. The edges are simply unordered pairs of distinct vertices $\mathcal{E}\subseteq \{\{u,v\}, u,v\in V, u\neq v\}$. We will denote $\{i,j\}\in\mathcal{E}$ if there is an edge between nodes $i$ and $j$, and similarly $\{i,j\}\notin\mathcal{E}$ if there is no such edge. There are no loops or double edges; the edges do not have directions, and there are no weights associated with nodes or edges. Alternatively, a graph $G$ can be described by a $N \times N$ symmetric matrix $A_{G}=(a_{ij})_{(i,j) \in \mathcal{V} \times   \mathcal{V}}$, where $a_{ij}=a_{ji}=1$ if $\{i,j\} \in \mathcal{E}$ is an edge of the graph $G$, $a_{ij}=a_{ji}=0$ if $\{i,j\} \notin \mathcal{E}$. Such a matrix is called the adjacency matrix of $G$. The symbol $\overline{G}$ defines the complement graph of $G$. More precisely, given a graph $G(\mathcal{V}, \mathcal{E})$ the complement graph is defined as a graph $\overline{G}(\mathcal{V}, \overline{\mathcal{E}})$, where $\overline{\mathcal{E}}=\{\{i,j\}: i,j\in \mathcal{V}, i\neq j$ and $ \{i,j\} \notin \mathcal{E}\}$. In words, a complement graph of a graph $G(\mathcal{V}, \mathcal{E})$ is the graph with the same vertex set as $G$, but with all edges that are not present in $G$.

\begin{figure}[ht]
    \centering
    \includegraphics[width=1\linewidth]{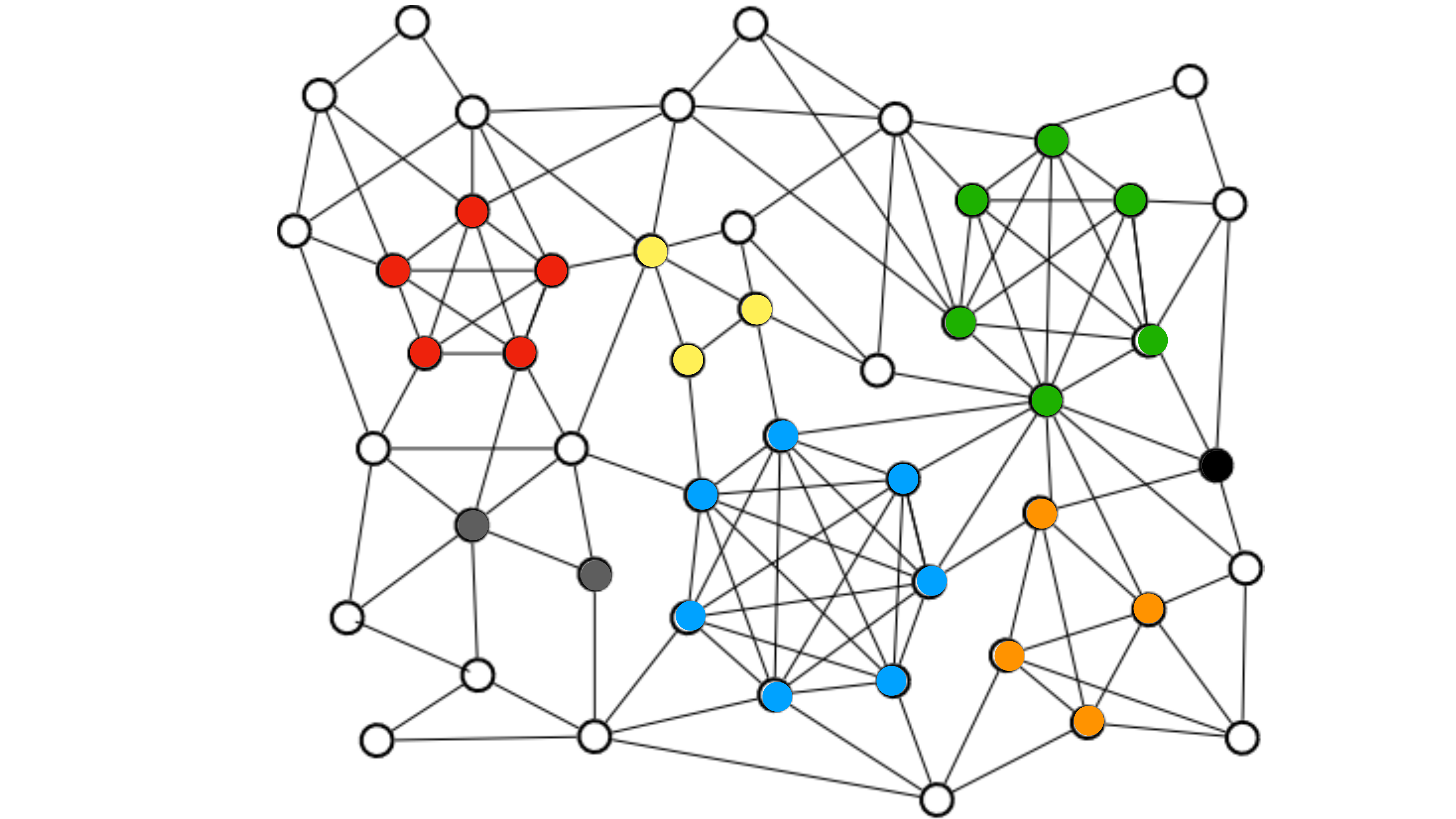}
    \caption{An illustrative example of cliques within an undirected
      graph, showcasing the interconnected nodes and the formation of
      tightly-knit groups. In black the size of a clique of order $1$,
      in grey of order $2$, in yellow of order $3$, in orange of order
      $4$, in red of order $5$, in green of order $6$, and in blue of
      order $7$.  This visualization emphasizes the concept of cliques
      as subsets of vertices within a graph, all of which are adjacent
      to each other, illustrating the density of connections that
      define these clusters. The maximum clique in the graph is the
      set of blue nodes. The blue, green, red, orange, and yellow cliques are also maximal; the black and gray cliques are not.
      }
    \label{fig:cliques}
\end{figure}

A complete graph $G(\mathcal{V}, \mathcal{E})$ can also be defined. In such a graph, each pair of distinct vertices is connected by a unique edge. In other words, a complete graph is a simple undirected graph where every pair of vertices is adjacent, i.e. $\forall i,j \in \mathcal{V} $ with $i \neq j$ we have $\{i,j\} \in \mathcal{E}$.  For a subset $S \subseteq \mathcal{V}$, we call $G(S)=G(S, \mathcal{E} \cap S \times S)$ the subgraph induced by $S$. We call the set $C$ a clique, if it induces a complete subgraph in $G(\mathcal{V}, \mathcal{E})$, that is $C \subseteq \mathcal{V}$ such that $G(C)$ is complete (see Fig.~\ref{fig:cliques}). We call the size of the clique the cardinality of $C$, i.e., $|C|$.

Given a node $v \in \mathcal{V}$, $N(v)$ denotes the neighborhood (or frontier) of $v$, i.e., the set of all nodes adjacent to $v$: $N(v) = \{j \in \mathcal{V}: \{v,j\}\in\mathcal{E}\}$.

We can define coloring of nodes by $k$ colors, that is, giving each node numbers from $1,\ldots,k$ as a label. Formally, the coloring of the nodes of $G$ can be described by a map $f : V \to \{1,\ldots, k\}$. We call a coloring proper or legal if two adjacent nodes receive different colors. That is if $i,j\in\mathcal{V}, i\neq j, \{i,j\}\in\mathcal{E}$ then $f(i)\neq f(j)$.

Closely related to cliques are the notions of independent sets and vertex covers in graphs. An independent set, or in other words a stable set, is a set of nodes that has no edge among them. A vertex cover is a set of nodes, such that every edge in the graph has at least one endpoint among these nodes. If given a graph $G(\mathcal{V}, \mathcal{E})$ and an independent set $I$, the set $K=\mathcal{V}\setminus I$ is a covering set -- presume that there would be an edge that has no endpoints in $K$, that would mean the endpoints of this edge are in $I$, which contradicts $I$ being an independent set. If given a graph $G(\mathcal{V}, \mathcal{E})$ and a covering set $K$, the set $I=\mathcal{V}\setminus K$ is an independent set -- as for every edge of the graph $G$ at least one endpoint lies in $K$, so there cannot be any edges between any two nodes of $I$.

\textit{We can now define some problems related to cliques.}

\begin{problem}
\label{kclq}
  Given a graph $G(\mathcal{V},\mathcal{E})$ and an
  integer $k$, decide if graph $G$ contains a clique of size $k$.
\end{problem}

This problem is called the $k$-clique problem. Sometimes it is also required to present a $k$-clique as well.

\begin{problem}
\label{mclq}
  Given a graph $G(\mathcal{V},\mathcal{E})$ find the
  maximum value of $k$, so that the graph $G$ has a clique of size
  $k$.
\end{problem}

This problem is called the Maximum Clique Problem (MCP). The size of a maximum clique is also called the clique number of the graph, and denoted in graph theory as $\omega(G)$. In other words, formally, $\omega(G)=\max\{|C|: C$ is a clique in $G\}$. Clearly, solving problem \ref{mclq} solves problem \ref{kclq} as well. Also, a small sequence of $k$-clique search would solve the MCP. In addition, one may also require that, apart from calculating the clique number, the program should present one of the maximum cliques as well. This is a slightly different problem, as in theory it would be possible to solve the first without actually presenting the clique in question. Although, to our knowledge all programs actually also find and present a maximum clique.

Compared to the maximum clique, one can speak about \textit{maximal} cliques. A maximal clique is a clique that cannot be extended; that is, no node can be added, so the resulting set of nodes is still a clique. A maximum clique is by definition always a maximal one, but
not vice versa.

\begin{problem}
\label{mclqall}
  Given a graph $G(\mathcal{V},\mathcal{E})$ list all
  maximum cliques in the graph $G$.
\end{problem}

This problem is clearly the most demanding one, and solving it, we got a solution to all previous ones.

There are equivalent problems to MCP. Finding a maximum clique in the complement of a graph ${G}(\mathcal{V}, \mathcal{E})$ is equivalent to finding a maximum set of vertices in $\overline{G}(\mathcal{V}, \overline{\mathcal{E})}$ whose elements are pairwise nonadjacent, thus a Maximum Independent Set (MIS) denoted by $\alpha(G)$. Given a Maximum Independent Set $I$ in a graph $G(\mathcal{V}, \mathcal{E})$ the set of sites obtained by the relation $\mathcal{V}\setminus I$ is a Minimum Vertex Cover set (MVC). Given that thus the MIS, MVC, and MCP are equivalent problems from the mathematical perspective. In the following sections, we will also discuss MIS and MVC results if we believe they are useful.

The MCP is well studied from the complexity viewpoint, with early complexity results on the problem being reviewed in Pardalos and Xue (1994)\cite{pardalos1994maximum}, Bomze et al. (1999)\cite{bomze1999maximum}, and Wu and Hao (2014)\cite{wu2015review}. However, during the last decade, a number of advances in the research of the computational complexity of the MCP appeared\cite{feldmann2020survey}. Although a full review of these theoretical results is clearly beyond the scope of this paper, we provide a brief summary of some main results.

\subsection{Hardness of the problem}

One can raise the question about how hard the proposed problem is from the point of view of practical computer science. First, we will examine the hardness of the exact solution of the problem. Second, we would like to discuss how effective a heuristic approach can be for this problem. Both approaches, the exact and the heuristic, are telling us that this problem is notoriously hard.

Evaluating the \textit{exact approach}, we would like to point to the historical fact that the decision formulation of MCP, the $k$-clique problem, problem \ref{kclq} as stated in the introduction, was first shown to be NP-complete in the seminal work of Karp\cite{karp1972reducibility}. Thus, its optimization variant, i.e., MCP, problem \ref{mclq}, where the goal is to find a clique of the largest possible size, is NP-hard. Given this result, the usual answer to the hardness of these problems is that they are exponential. But there are several factors here.

First, that this problem class, namely the worst-case running time of these problems would be exponential is only a hypothesis, named the Exponential Time Hypothesis (ETH)\cite{ETH}. So even in the case of P$\neq$NP, it may turn out that the complexity of these problems are still sub-exponential.

Second, obviously, different kinds of problems or even problem instances can be very easy or very hard. As even the ETH is only stating exponential running time for the worst case. So consequently, for algorithm engineering, it is of utmost importance to have a good set of benchmarks for testing. A usual metric of the graph that can sometimes, but not always, point to the hardness is the density of the graph. As shown in the outstanding paper of Scottish researchers densities around 80\%--95\% are usually the hardest, and this research is also backed up with the experimental results of different clique search algorithms\cite{hardnessMcCreesh}.

Third, the problems stated above are inherently different. Take, for example, the $k$-clique problem, problem \ref{kclq}. On one hand, if a heuristic algorithm finds a clique of size $k$ one has already solved the problem. Also, if someone is using a heuristic way to find an upper bound, and that bound is less than $k$, this also solves the problem -- there cannot be a $k$-clique in the graph. On the other hand if there is no $k$-clique present in the graph, and one cannot find an upper bound to prove it, then solving this problem will be hard.

For the MCP, finding a big clique by heuristics does not solve the problem alone, only if there is also an upper bound such that the value of the bound is equal to the size of the found clique. In all other cases, one needs to find the maximum clique \textit{and} prove that there is no bigger clique present -- and this is hard.

As for the problem \ref{mclqall}, listing all maximum cliques, there seems to be no shortcut, no heuristics to help solution, the problem is always hard.

Finally, there is a curious and sometimes confusing difference between MVC and MCP, althought we stated that they are the same from a mathematical point of view. MVC is famous for being fixed parameter tractable (FPT), while MCP is \textit{not}, see ref.\cite{parametrized99,parametrized13}. (Formally, the $k$-clique problem is in the W$[1]$ complexity class. The reader can find the more technical definition on the W hierarchy and the fixed parameter tractability in  ref.\cite{parametrized99,parametrized13}.) Recall, that FPT means that the problem can be solved in time $f(k)\cdot {|x|}^{O(1)}$ for some computable function $f$. In our case, $x$ is the size of the graph, that is $|\mathcal{V}|$, and $k$ would be the size of a minimum vertex cover set, or the size of a maximum clique. Though if $K$ is a covering set, the set $I=\mathcal{V}\setminus K$ is an independent set, and $C=\mathcal{V}\setminus K$ is a clique in the complement graph $\overline{G}$. That is $|K|=|\mathcal{V}|-|C|$. Which means, that if MVC is small, that is $k=|K|, k\ll |V|$, the parameterized problem is dominated by the $f(k)$ function. The corresponding clique problem, though, has a solution where almost all nodes are in the clique. In simple words MVC is FPT for small set sizes, while the clique problem is FPT for clique number close to the size of the set of all nodes -- with the parameter value $(|\mathcal{V}|-k)$. For small clique numbers, the problem is clearly not FPT, that is why the clique problem is not in the FPT class (by the parameter of the clique size).

Given that the exact solution of the MCP is computationally hard one would decide to use \textit{heuristic approach} in order to find a suitably good solution instead of a proven one. That is to find a big enough (maximal) clique, and hope that the size of this clique would be close to the size of a maximum clique. This approach is driven by the fact that some real-world problems require finding a big enough clique and do not need a proof of optimality. But this approach is also hard by itself, as it is proven, that even approximating the clique number is hard\cite{haastad1999clique}. This means that, opposed to other problems, such as the Knapsack Problem or the metric Traveling Salesman Problem, where there are effective constant-factor approximation algorithms, the approximation of MCP cannot be such effective.

The state-of-the-art of the best-known polynomial-time approximation algorithm is due to Feige\cite{feige2004approximating}, which yields an approximation ratio of $O(N(\log \log N)^2/(\log N)^3)$. The approximation ratio is the ratio between the size of the maximum clique and the size of the solution output by the algorithm. In Engebretsen et al.\cite{engebretsen2003towards}, however, the authors showed that the MCP is not approximable within a factor of $N/2^{O(\log N)/\sqrt{\log \log N}}$ under the assumption that NP$\subseteq$ ZPTIME($2^{O(\log N)(\log \log N)^{3/2}}$). An improved result shows that the MCP is not approximable within $N^{1-\epsilon}$, with $\epsilon > 0$, unless P=NP\cite{zuckerman2006linear}.

In Chitnis et al.\cite{chitnis2013fixed} it is established that unless NP$\subseteq$ SUBEXP, for any $0<\delta<1$ there exists a constant, that depends by $\delta$, i.e., $F(\delta)>0$, such that the MCP has no Fixed-Parameter Tractable (FPT) optimum approximation with ratio $\rho(OPT) =OPT^{1-\delta}$ in $2^{{OPT}^{F}} poly(|\mathcal{V}|)$. Finally, in 2020 the authors in Chalermsook et al.\cite{doi:10.1137/18M1166869} showed that there is no nontrivial FPT approximation algorithm for MCP. More precisely, there is no $o(OPT)$-FPT-approximation algorithm for MCP (for an updated summary on MIS, we refer to Marino et al.\cite{marino2023large}).

Research in this direction came up with an interesting approach, which in the last decade has obtained attention from the scientific community and has been barely mentioned in the reviews\cite{bomze1999maximum, wu2015review}: the hidden clique problem (HCP), or planted clique model. It has received a great deal of attention and has become a hallmark of a research area that is now called the study of statistical-computational gaps. These are the regimes where a signal is detectable or recoverable yet believed to be intractable for polynomial-time algorithms. A common strategy to provide evidence for such a gap is to prove that powerful classes of efficient algorithms are unable to solve the planted problem in the (conjecturally) hard regime.

Several authors have proposed that the search for powerful, effective clique-finding algorithms could be expressed as a challenge, perhaps to attract the widest set of challengers to the problem. Mark Jerrum, in his 1992 paper\cite{jerrum1992large} sets out several of these. His paper shows that a restricted version of stochastic search is unlikely to reach a maximum clique, and also introduces the additional problem of finding an artificially hidden clique.  A hidden clique or \textit{planted solution}, is just what it sounds like, a single subgraph of sites, with $|C_{HC}| > |C_{\max}|$, so that it can be distinguished, for which all the missing bonds among those sites have been restored.  A series of papers\cite{alon1998finding,dekel2014finding} show the existence of polynomial algorithms that are able to compute a hidden clique of size $|C_{HC}|$ proportional to $N^{0.5}$. The state-of-the-art for the HCP is given by Deshpande et al.\cite{deshpande2015finding}. They developed a rigorous analysis that is asymptotically exact as $N \to \infty$, and they prove that their algorithm is able to find hidden cliques of size $|C_{HC}| \ge \sqrt{N/e}$ with high probability.  

For computing lower bounds, a very useful method has been established in recent years. The name of such method is sum of squares (SOS) semidefinite hierarchy. SOS algorithms are particularly attractive targets for lower bounds because of their broad applicability and strong guarantees\cite{Hopkins2017}.

The sum-of-squares (SOS) hierarchy, also known as the Lasserre hierarchy\cite{Lasserre2001}, is a hierarchy of convex relaxations of increasing power and increasing computational cost. For each integer $r\ge 1$, the corresponding convex relaxation is known as the $r$-th round of the SOS hierarchy.  The first round ($r=1$) coincides with basic semidefinite programming, or to sum-of-squares optimization over polynomials of degree at most $2$. Passing from level $r$ to $r+1$ adds the program with variables and constraints so that certificates may involve polynomials up to degree $2(r+1)$. In ref.\cite{meka2015sum}, the authors proved the first average case lower bound for HCP. They showed that for $r=1$, SOS can find solution up to $|C_{HC}|=\Theta(\sqrt{N})$. Deshpande et al.\cite{deshpande2015improved}, instead, considered the $r=4$ SOS relaxation. This is the first level at which the SOS hierarchy diﬀers substantially from the baseline spectral algorithm of Alon, Krivelevich and Sudakov\cite{alon1998finding}. The authors in ref.\cite{deshpande2015improved} proved that SOS fails unless $|C_{HC}|\geq c N^{1/3}/\log N$. Their proof uses the moment method to bound the spectrum of a certain random association scheme, i.e. a symmetric random matrix whose rows and columns are indexed by the edges of an Erd\H{o}s-R\'enyi random graph.

A full review of these theoretical results is clearly beyond the scope of this manuscript; however, one can refer to ref.\cite{pang2021sos, jones2022sum, jones2023sum, 10353221} and references therein for a more accurate description of the SOS methodology.

\subsection{Mathematical programming formulation of MCP}

One obvious way to solve the MCP is to use a mathematical programming solver, such as Integer Linear Programming (ILP) solver.

The formulation of the MCP has been the subject of many studies, which offer valuable insights into the problem and its theoretical and practical implications. Reviewing these studies is important for gaining a comprehensive understanding of existing MCP formulations and discovering new results. Bomze et al. (1999)\cite{bomze1999maximum}, and Pardalos and Xue (1994)\cite{pardalos1994maximum} provide an in-depth analysis of existing formulations, while we will focus on some common and recently developed formulations.

Following ref.\cite{bomze1999maximum, wu2015review, Hosseinian2017}, the MCP can be formulated in the following manner: let's $x_i \in\{0,1\}, i=1,...,N$, we consider to find the $\max \sum_{i=1}^N x_i$ under the constraint that $x_i+x_j \leq 1, \forall \{i,j\} \in \overline{\mathcal{E}}$, which is usually called the edge formulation.

In this simple formulation any feasible solution of MCP on a graph $G$ is obtained by the set of vertices $C$ such that a vertex $i \in C$ if $x_i=1$ and $x_i=0$ otherwise. Nemhauser and Trotter\cite{nemhauser1975vertex,nemhauser1974properties} showed that if a variable $x_i = 1$ holds for an optimal solution to the linear relaxation of the above formulation, then $x_i=1$ holds for at least one optimal solution to the integer formulation. Clearly, this result can be used by an algorithm to reduce the search space explored when seeking an optimal clique. 

The same result can be obtained using an alternative formulation based on independent sets. By denoting with $S$  the set of all maximal independent sets in $G$, any clique $C$ in $G$ can contain no more than a single vertex from such a set. Mathematically, one wants to maximize $\sum_{i=1}^N x_i$, with $x_i \in\{0,1\}, i=1,...,N$,  under the constraint $\sum_{i\in s} x_i \leq 1, \forall s \in S$. This formulation offers the benefit of a minimal discrepancy between an optimal solution and its linear relaxation. However, cataloging all independent sets within a generic graph presents a challenge, as the number of independent sets increases exponentially, corresponding to the size of the graph\cite{Moon1965}. Thus, applying the relaxed version of the independent set formulation is a nontrivial task, and probably calls for a column generation approach, which is out of our scope.

The reformulation of the problem of maximum weighted clique is straightforward using any previously detailed methods. The only difference is that one needs to add weights to the objective function. The edge-weighted version of the problem is far more challenging. For sparse graphs, using variables representing edges can be a viable solution\cite{Gouveia2015}. For dense graphs other approaches may be needed such as using quadratic programming\cite{Hosseinian2018}. The quadratic reformulation as an alternative technique can be used for the node weighted version as well, and it can, in some cases, generate good computational results\cite{shor1990dual}. For connections between the MCP and a certain standard quadratic programming problem, we refer to ref.\cite{motzkin1965maxima}.

Remarkable discretized formulations of MCP were presented in Martins\cite{martins2010extended}. By using additional and extra indexed variables and constraints, the formulations depend on the range of variation of an interval containing the clique number of the graph. Therefore, tight lower and upper bounds may strongly benefit the dimension of the formulations. Lower bounds identify cliques that an efficient algorithm can find, while upper bounds are on the actual maximum clique, not just on the size an algorithm can find.  A simple application of the probabilistic method can reveal with a high degree of certainty that the maximum clique, in an Erd\H{o}s-R\'enyi graph $G(|\mathcal{V}|=N,p)$ with probability $p$ that two vertices are linked each other, approximates to $d(N)=2 \log_{1/p}N - 2 \log_{1/p}\log_{1/p}N + 2\log_{1/p} \frac{e}{2} +1 + O(1)$ for large $N$\cite{bollobas1976cliques}. Matula, instead, drew attention to what is now termed a concentration result for the MCP.  As $N \to \infty$ the sizes of the largest cliques $C_{\max}$ that will occur are concentrated on just two values, the integers immediately below and above $|C_{\max}|=d(N)$.  To do this, he used the second moment of the distribution of the numbers of cliques of a certain size to bound the fraction of graphs with no such cliques and sharpened the result of Matula\cite{matula1976largest} by computing a weighted second moment. Indeed, Markov's inequality provides upper bounds, and Chebyshev's inequality provides lower bounds on the existence of such cliques\cite{matula1970complete,matula1972employee,matula1976largest}.  The fraction of graphs $G(N,p)$ with maximum clique size $|C_{\max}|$, is bounded as follows: \begin{equation}\label{matula}
\left(\sum_{j=\max\{0, 2k-N\}}^k \frac{{N-k \choose k-j}{k \choose j}}{{N \choose k}} p^{-{j \choose 2}}  \right)^{-1}\leq P(|C_{\max}|\geq k) \leq {N \choose k} p^{k \choose 2}. \end{equation}

Recently, in Belachew et al.\cite{belachew2017solving}, it was formulated a very convenient continuous characterization of the maximum clique problem based on the symmetric rank-one non-negative approximation of a given matrix and building a one-to-one correspondence between stationary points and cliques of a given graph $G$. In particular, the authors showed that the local (resp. global) minima of the continuous problem correspond to the maximal (resp. maximum) cliques of the given graph $G$.

The last formulation of the MCP is the one best suited for quantum computers in the form of a Quadratic Unconstrained Binary Optimization (QUBO) problem. To do that, we first take the trivial mapping from MCP to MIS on the dual graph. Then, given a graph $G = (V, E)$, in order to find the MIS, we need to assign a binary variable $x_i$ to each one of the $i$ vertices, which is $1$ if the vertex belongs to the MIS and $0$ otherwise. Thus, this problem becomes:
\begin{equation}
    \max_{x_i \in \{0,1\}} \sum_{i=1}^{N} x_i
\end{equation}
subjected to the constraint
\begin{equation}
    \sum_{(i,j) \in E} x_i x_j = 0,
\end{equation}
The equivalent QUBO representation of this problem, expressed in Hamiltonian form, is
\begin{equation}\label{eq:QUBO_Hamiltonian}
    H = -A \sum_{i=1}^{N} x_i + B \sum_{(i,j) \in E} x_i x_j,
\end{equation}
where the coefficients (penalties) $A$ and $B$ can be chosen appropriately as discussed in ref.\cite{lucas2014ising,chapuis2017finding}.

\subsection{Benchmark Collections}

As we already pointed out, algorithm design is always going along with the question of testing. Historically, algorithms for MCP were tested on random graphs based on the Erd\H{o}s-R\'enyi model. Such a graph, $G(N,p)$, is a graph of $N$ nodes. For each pair of nodes there is an edge with probability $p$ included independently from every other edge. This graph class is easy to generate and so test algorithms against it. But these graphs are far from perfect for testing, as they lack any structure.

In evaluating the performance of algorithms designed for the MCP many benchmarks appeared in the academic literature. In the following, we list the most commonly used benchmarks for testing algorithms.

\subsubsection{DIMACS}
The DIMACS benchmark from 1996 was part of the second DIMACS Implementation Challenge\cite{johnson1996cliques}. This conference focused on problems related to Clique, Satisfiability, and Graph Coloring. This challenge led to the creation of a set of standard benchmark graphs that have been extensively used to evaluate algorithms for these problems. It is composed of $80$ graphs, of different order and size, as presented in Jin et al.\cite{jin2015general}. Note, that there are artificially constructed graphs that thus have a known clique number, but still cannot be solved by a recent exact solver. Such are for example the graphs coming from coding theory like some of the hamming graphs and johnson graphs\cite{hamming10-4,johnson32-2-3}.

The DIMACS benchmark encompasses a diverse array of practical challeng es, including those derived from coding theory, fault diagnosis, and Keller's conjecture, among others. This collection not only features graphs generated through random processes but also includes graphs specifically designed to obscure the maximum clique by integrating vertices of lower degrees. The scale of these DIMACS instances is quite broad, with some comprising fewer than $50$ vertices and $10^3$ edges, while others expand to over $3300$ vertices and $5$ million edges. Based on documented findings, the majority of the $80$ instances in the DIMACS benchmark have had their maximum cliques identified. However, some exceptions persist. Today, experimental results obtained by an exact algorithm show that the number of instances solved to proven optimality is $77$\cite{LI20171}. A list of best-known solutions in literature for the DIMACS benchmark is given in Table \ref{table:dimacs}. In this table, and also in the following ones, we indicated the name of the file, the number of nodes ($|\mathcal{V}|$), the number of edges ($|\mathcal{E}|$), and the size of the maximum clique ($\omega(G)$).

\begin{table*}[h!]\label{IMACS}
\centering
\begin{minipage}{.5\linewidth}
\centering
\begin{tabular}{ |l|l|l|l| }
\hline
Name & $|\mathcal{V}|$ & $|\mathcal{E}|$ & $\omega(G)$ \\
\hline
brock200\_1 & 200 & 14834 & 21$^*$\\
brock200\_2 & 200 & 9876 & 12$^*$ \\
brock200\_3 & 200 & 12048 & 15$^*$ \\
brock200\_4 & 200 & 13089 & 17$^*$ \\
brock400\_1 & 400 & 59723 & 27$^*$ \\
brock400\_2 & 400 & 59786 & 29$^*$ \\
brock400\_3 & 400 & 59681 & 31$^*$ \\
brock400\_4 & 400 & 59765 & 33$^*$ \\
brock800\_1 & 800 & 207505 & 23$^*$ \\
brock800\_2 & 800 & 208166 & 24$^*$ \\
brock800\_3 & 800 & 207333 & 25$^*$ \\
brock800\_4 & 800 & 207643 & 26$^*$ \\
C125.9 & 125 & 6963 & 34$^*$ \\
C250.9 & 250 & 27984 & 44$^*$ \\
C500.9 & 500 & 112332 & $\geq 57$ \\
C1000.9 & 1000 & 450079 & $\geq 68$ \\
C2000.5 & 2000 & 999836 & 16$^*$ \\
C2000.9 & 2000 & 1799532 & $\geq 80$ \\
C4000.5 & 4000 & 4000268 & 18$^*$ \\
DSJC500.5 & 500 & 125248 & 13$^*$ \\
DSJC1000.5 & 1000 & 499652 & 15$^*$\\
keller4 & 171 & 9435 & 11$^*$ \\
keller5 & 776 & 225990 & 27$^*$ \\
keller6 & 3361 & 4619898 & 59$^*$ \\
MANN\_a9 & 45 & 918 & 16$^*$ \\
MANN\_a27 & 378 & 70551 & 126$^*$ \\
MANN\_a45 & 1035 & 533115 & 345$^*$ \\
MANN\_a81 & 3321 & 5506380 & 1100$^*$ \\
hamming6-2 & 64 & 1824 & 32$^*$ \\
hamming6-4 & 64 & 704 & 4$^*$ \\
hamming8-2 & 256 & 31616 & 128$^*$ \\
hamming8-4 & 256 & 20864 & 16$^*$ \\
hamming10-2 & 1024 & 518656 & 512$^*$ \\
hamming10-4 & 1024 & 434176 & 40$^*$\\
gen200-p0.9-44 & 200 & 17910 & 44$^*$ \\
gen200-p0.9-55 & 200 & 17910 & 55$^*$ \\
gen400-p0.9-55 & 400 & 71820 & 55$^*$ \\
gen400-p0.9-65 & 400 & 71820 & 65$^*$ \\
gen400-p0.9-75 & 400 & 71820 & 75$^*$\\
c-fat200-1 & 200 & 1534 & 12$^*$ \\
\hline
\end{tabular}
\end{minipage}%
\begin{minipage}{.5\linewidth}
\centering
\begin{tabular}{ |l|l|l|l| }
\hline
Name & $|\mathcal{V}|$ & $|\mathcal{E}|$ & $\omega(G)$ \\
\hline
c-fat200-2 & 200 & 3235 & 24$^*$ \\
c-fat200-5 & 200 & 8473 & 58$^*$ \\
c-fat500-1 & 500 & 4459 & 14$^*$ \\
c-fat500-2 & 500 & 9139 & 26$^*$ \\
c-fat500-5 & 500 & 23191 & 64$^*$ \\
c-fat500-10 & 500 & 46627 & 126$^*$ \\
johnson8-2-4 & 28 & 210 & 4$^*$ \\
johnson8-4-4 & 70 & 1855 & 14$^*$ \\
johnson16-2-4 & 120 & 5460 & 8$^*$ \\
johnson32-2-4 & 496 & 107880 & 16$^*$ \\
p\_hat300-1 & 300 & 10933 & 8$^*$ \\
p\_hat300-2 & 300 & 21928 & 25$^*$ \\
p\_hat300-3 & 300 & 33390 & 36$^*$ \\
p\_hat500-1 & 500 & 31569 & 9$^*$ \\
p\_hat500-2 & 500 & 62946 & 36$^*$ \\
p\_hat500-3 & 500 & 93800 & 50$^*$ \\
p\_hat700-1 & 700 & 60999 & 11$^*$ \\
p\_hat700-2 & 700 & 121728 & 44$^*$ \\
p\_hat700-3 & 700 & 183010 & 62$^*$ \\
p\_hat1000-1 & 1000 & 122253 & 10$^*$ \\
p\_hat1000-2 & 1000 & 244799 & 46$^*$ \\
p\_hat1000-3 & 1000 & 371746 & 68$^*$ \\
p\_hat1500-1 & 1500 & 284923 & 12$^*$ \\
p\_hat1500-2 & 1500 & 568960 & 65$^*$ \\
p\_hat1500-3 & 1500 & 847244 & 94$^*$ \\
san200\_0.7\_1 & 200 & 13930 & 30$^*$ \\
san200\_0.7\_2 & 200 & 13930 & 18$^*$ \\
san200\_0.9\_1 & 200 & 17910 & 70$^*$ \\
san200\_0.9\_2 & 200 & 17910 & 60$^*$ \\
san200\_0.9\_3 & 200 & 17910 & 44$^*$ \\
san400\_0.5\_1 & 400 & 39900 & 13$^*$ \\
san400\_0.7\_1 & 400 & 55860 & 40$^*$ \\
san400\_0.7\_2 & 400 & 55860 & 30$^*$ \\
san400\_0.7\_3 & 400 & 55860 & 22$^*$ \\
san400\_0.9\_1 & 400 & 71820 & 100$^*$ \\
san1000 & 1000 & 250500 & 15$^*$ \\
sanr200\_0.7 & 200 & 13868 & 18$^*$ \\
sanr200\_0.9 & 200 & 17863 & 42$^*$ \\
sanr400\_0.5 & 400 & 39984 & 13$^*$ \\
sanr400\_0.7 & 400 & 55869 & 21$^*$ \\
\hline
\end{tabular}
\end{minipage}
\caption{DIMACS best known results in literature for $\omega(G)$ (optimal
values are marked with '$^*$').}
\label{table:dimacs}
\end{table*}

\subsubsection{BHOSLIB}

The BHOSLIB benchmark is a collection of graph instances specifically designed for testing several clique algorithms. The BHOSLIB benchmark is composed of $41$ graphs with hidden optimal solutions\cite{xu2007random}. The instances in BHOSLIB are known for being particularly challenging. They are translated from hard random SAT instances generated at the exact phase transition of the RB model\cite{xu2005simple}. The size of these instances varies, starting from graphs with $450$ vertices and $83198$ edges, and scaling up to graphs with $4000$ vertices and $7$ million edges. Although designed for a heuristic approach these graphs were used as test graphs for different exact algorithms. it was found that only a limited number of such algorithms were able to effectively address them. On the other hand, several recent heuristic algorithms can attain the known optimal solutions for these instances with no particular difficulty\cite{jin2017algorithms,pan2023improved,xiao2021efficient,yin2023solving,cai2017finding}. For comprehensive information on this benchmark and to access the downloadable instances, please visit the following website: \href{https://iridia.ulb.ac.be/~fmascia/maximum_clique/BHOSLIB-benchmark#detfrb100-40}{BHOSLIB Benchmark}. For the sake of simplicity, here we report the list of the hidden solution size in Table \ref{table:bhoslib}

\begin{table*}[h!]
\centering
\begin{minipage}{.5\linewidth}
\centering
\begin{tabular}{ |l|l|l|l| }
\hline
Name & $|\mathcal{V}|$ & $|\mathcal{E}|$ & $\omega(G)$ \\
\hline
frb30-15-1 & 400 & 83198 & 30$^*$ \\
frb30-15-2 & 400 & 83151 & 30$^*$ \\
frb30-15-3 & 400 & 83216 & 30$^*$ \\
frb30-15-4 & 400 & 83194 & 30$^*$ \\
frb30-15-5 & 400 & 83231 & 30$^*$ \\
frb35-17-1 & 595 & 148859 & 35$^*$ \\
frb35-17-2 & 595 & 148868 & 35$^*$ \\
frb35-17-3 & 595 & 148784 & 35$^*$ \\
frb35-17-4 & 595 & 148873 & 35$^*$ \\
frb35-17-5 & 595 & 148572 & 35$^*$ \\
frb40-19-1 & 760 & 247106 & 40$^*$ \\
frb40-19-2 & 760 & 247157 & 40$^*$ \\
frb40-19-3 & 760 & 247325 & 40$^*$ \\
frb40-19-4 & 760 & 246815 & 40$^*$ \\
frb40-19-5 & 760 & 246801 & 40$^*$ \\
frb45-21-1 & 945 & 386854 & 45$^*$ \\
frb45-21-2 & 945 & 387416 & 45$^*$ \\
frb45-21-3 & 945 & 387795 & 45$^*$ \\
frb45-21-4 & 945 & 387491 & 45$^*$ \\
frb45-21-5 & 945 & 387461 & 45$^*$ \\
\hline
\end{tabular}
\end{minipage}%
\begin{minipage}{.5\linewidth}
\centering
\begin{tabular}{ |l|l|l|l| }
\hline
Name & $|\mathcal{V}|$ & $|\mathcal{E}|$ & $\omega(G)$ \\
\hline
frb50-23-1 & 1150 & 580603 & 50$^*$ \\
frb50-23-2 & 1150 & 579824 & 50$^*$ \\
frb50-23-3 & 1150 & 579607 & 50$^*$ \\
frb50-23-4 & 1150 & 580417 & 50$^*$ \\
frb50-23-5 & 1150 & 580640 & 50$^*$ \\
frb53-24-1 & 1272 & 714129 & 53$^*$ \\
frb53-24-2 & 1272 & 714067 & 53$^*$ \\
frb53-24-3 & 1272 & 714229 & 53$^*$ \\
frb53-24-4 & 1272 & 714048 & 53$^*$ \\
frb53-24-5 & 1272 & 714130 & 53$^*$ \\
frb56-25-1 & 1400 & 869624 & 56$^*$ \\
frb56-25-2 & 1400 & 869899 & 56$^*$ \\
frb56-25-3 & 1400 & 869921 & 56$^*$ \\
frb56-25-4 & 1400 & 869262 & 56$^*$ \\
frb56-25-5 & 1400 & 869699 & 56$^*$ \\
frb59-26-1 & 1534 & 1049256 & 59$^*$ \\
frb59-26-2 & 1534 & 1049648 & 59$^*$ \\
frb59-26-3 & 1534 & 1049729 & 59$^*$ \\
frb59-26-4 & 1534 & 1048800 & 59$^*$ \\
frb59-26-5 & 1534 & 1049829 & 59$^*$ \\
\hline
\end{tabular}
\end{minipage}
\caption{BHOSLIB benchmark with size of planted clique of size $\omega(G)$ (optimal
values are marked with '$^*$').}
\label{table:bhoslib}
\end{table*}

\subsubsection{EVIL}
The Evil benchmark\cite{szabo2019benchmark} is notable for its ability to create challenges of varying difficulty levels for maximum clique search algorithms. This adaptability is crucial because many existing benchmark problems are either too simple for modern solvers or excessively challenging. The Evil benchmark addresses this gap by offering a range of tests that can be fine-tuned for different degrees of difficulty. The benchmark's design intentionally widens the gap between the chromatic number and the clique number of the graphs, 
while using the hidden clique technique in a (partially) random graph, making it a valuable tool for algorithm developers and researchers. This aspect of the benchmark allows for a more comprehensive assessment of the algorithms' capabilities, especially in scenarios where traditional benchmarks might not provide a sufficient challenge or might not accurately represent the complexity of real-world problems. For the presented work we extended the original set of graphs to conclude some more challenging instances for the novel algorithms.
The size of the new set of instances varies, starting from graphs with $120$ vertices and $6595$ edges, and scaling up to graphs with $750$ vertices and $270928$ edges. 
In summary, the Evil benchmark for the Maximum Clique Problem offers a versatile and challenging set of small test instances, allowing for a thorough and nuanced evaluation of maximum clique search algorithms. This benchmark stands out for its ability to provide a wide range of difficulties, making it a valuable resource in the field of algorithmic research and development. The benchmark is composed of 40 graphs. For the sake of simplicity, here we report them in Table \ref{table:evil} To download the instances please visit: \href{https://github.com/zbogdan/evil2}{Evil Benchmark graphs}

\begin{table*}[h!]
\centering
\begin{minipage}{.5\linewidth}
\centering
\begin{tabular}{ |l|l|l|l| }
\hline
Name & $|\mathcal{V}|$ & $|\mathcal{E}|$ & $\omega(G)$ \\
\hline
chv12x10 & 120 & 6595 & 20$^*$ \\
myc5x24 & 120 & 6904 & 48$^*$ \\
myc11x11 & 121 & 6752 & 22$^*$\\
s3m25x5 & 125 & 6877 & 20$^*$\\
myc23x6 & 138 & 8211 & 12$^*$\\
myc5x30 & 150 & 10837 & 60$^*$\\
s3m25x6 & 150 & 10073 & 24$^*$\\
myc11x14 & 154 & 11080 & 28$^*$\\
chv12x15 & 180 & 15166 & 30$^*$\\
myc5x36 & 180 & 15671 & 72$^*$\\
myc23x8 & 184 & 15072 & 16$^*$\\
myc11x17 & 187 & 16490 & 34$^*$\\
s3m25x8 & 200 & 18350 & 32$^*$\\
myc5x42 & 210 & 21404 & 84$^*$\\
myc11x20 & 220 & 22960 & 40$^*$\\
myc23x10 & 230 & 24072 & 20$^*$\\
chv12x20 & 240 & 27328 & 40$^*$\\
myc5x48 & 240 & 27962 & 96$^*$\\
s3m25x10 & 250 & 29075 & 40$^*$\\
myc11x23 & 253 & 30422 & 46$^*$\\
\hline
\end{tabular}
\end{minipage}%
\begin{minipage}{.5\linewidth}
\centering
\begin{tabular}{ |l|l|l|l| }
\hline
Name & $|\mathcal{V}|$ & $|\mathcal{E}|$ & $\omega(G)$ \\
\hline
chv12x30 & 360 & 62211 & 60$^*$ \\
chv12x40 & 480 & 111059 & 80$^*$ \\
chv12x60 & 720 & 251269 & 120$^*$ \\
myc5x60 & 300 & 43659 & 120$^*$ \\
myc5x80 & 400 & 77893 & 160$^*$ \\
myc5x100 & 500 & 121781 & 200$^*$ \\
myc5x120 & 600 & 175458 & 240$^*$ \\
myc5x140 & 700 & 239044 & 280$^*$ \\
myc11x30 & 330 & 52214 & 60$^*$ \\
myc11x40 & 440 & 93327 & 80$^*$ \\
myc11x50 & 550 & 146280 & 100$^*$ \\
myc11x60 & 660 & 210997 & 120$^*$ \\
myc23x15 & 345 & 55464 & 30$^*$ \\
myc23x20 & 460 & 99896 & 40$^*$ \\
myc23x25 & 575 & 157359 & 50$^*$ \\
myc23x30 & 690 & 227595 & 60$^*$ \\
s3m25x15 & 375 & 66480 & 60$^*$ \\
s3m25x20 & 500 & 119354 & 80$^*$ \\
s3m25x25 & 625 & 187499 & 100$^*$ \\
s3m25x30 & 750 & 270928 & 120$^*$ \\
\hline
\end{tabular}
\end{minipage}
\caption{Evil benchmarks with size of planted clique of size $\omega(G)$ (optimal values are marked with '$^*$').}
\label{table:evil}
\end{table*}

\subsubsection{Benchmarks From Coding Theory}

Other instances are also widely used in MCP algorithm development community, such as instances arising from coding theory\cite{sloan2000,deletion11is172}. Here, the question is to find a maximum size code under some channel restrictions, such as Single-Deletion-Correcting Codes, Two-Deletion-Correcting Codes, Codes For Correcting a Single Transposition, etc. The task is to find the maximum independent set, so in the present work, we list the properties of the complement graphs for the MCP. We list these instances in Table \ref{table:coding}, some instances have optimal value proven, but for some instances, we have only lower and upper bounds. These instances greatly vary in size and hardness, some are still unsolved. They are considered as very challenging problems.

\begin{table*}[h!]\label{code}
\centering
\begin{minipage}{.5\linewidth}
\centering
\begin{tabular}{ |l|l|l|l| }
\hline
Name & $|\mathcal{V}|$ & $|\overline{\mathcal{E}}|$ & $\omega(G)$ \\
\hline
1dc.64 & 64 &  1473 & 10$^*$ \\
1dc.128 & 128 &  6657 & 16$^*$ \\
1dc.256 & 256 &  28801 & 30$^*$ \\
1dc.512 & 512 &  121089 & 52$^*$ \\
1dc.1024 & 1024 &  499713 & 94$^*$ \\
1dc.2048 & 2048 &  2037761 & 172$^*$ \\
2dc.128 & 128 &  2955 & 5$^*$ \\
2dc.256 & 256 &  15457 & 7$^*$ \\
2dc.512 & 512 &  75921 & 11$^*$ \\
2dc.1024 & 1024 &  354614 & 16$^*$ \\
2dc.2048 & 2048 &  1591677 & 24$^*$ \\
1tc.8 & 8 &  22 & 4$^*$ \\
1tc.16 & 16 &  98 & 8$^*$ \\
1tc.32 & 32 & 428 & 12$^*$ \\
1tc.64 & 64 &  1824 & 20$^*$ \\
1tc.128 & 128 &  7616 & 38$^*$ \\
\hline
\end{tabular}
\end{minipage}%
\begin{minipage}{.5\linewidth}
\centering
\begin{tabular}{ |l|l|l|l| }
\hline
Name & $|\mathcal{V}|$ & $|\overline{\mathcal{E}}|$ & $\omega(G)$ \\
\hline
1tc.256 & 256  &  31328 & 63$^*$ \\
1tc.512 & 512 &  127552 & 110$^*$ \\
1tc.1024 & 1024 & 515840 & 196$^*$ \\
1tc.2048 & 2048 & 2077184 & 352$^*$ \\
1et.64 & 64 & 1752 &18$^*$ \\
1et.128 & 128 & 7456 &28$^*$ \\
1et.256 & 256 & 30976 & 50$^*$ \\
1et.512 & 512 & 126784 & 100$^*$ \\
1et.1024 & 1024  & 514176 & 171$^*$ \\
1et.2048 & 2048 & 2073600 & 316$^*$ \\
1zc.128 & 128 & 5888 & 18$^*$ \\
1zc.256 & 256 & 27008 & 36$^*$ \\
1zc.512 & 512 & 116992 & 62$^*$ \\
1zc.1024 & 1024 & 490496 & 112--117 \\
1zc.2048 & 2048 & 2017280 & 198--210 \\
1zc.4096 & 4096 & 8202240 & 379--410\\
\hline
\end{tabular}
\end{minipage}
\caption{Graphs arising from coding theory (optimal
values are marked with '$^*$').}
\label{table:coding}
\end{table*}

\subsubsection{A Miscellaneous Benchmark List}

In 2019, the 4th Parameterized Algorithms and Computational Experiments (PACE) challenge\cite{dzulfikar2019pace} notably featured a track on the Minimum Vertex Cover (MVC) problem. The challenge, organized by the PACE community, provided a significant contribution to the computational research community in the form of benchmarks. These benchmarks are particularly valuable. They offer a comprehensive set of instances that can be repurposed for testing and evaluating algorithms for the MCP. Specifically, the PACE 2019 challenge's Vertex Cover instances encompass a diverse range of scenarios, catering to the assessment of various algorithmic strategies. This diversity stems from the inclusion of instances originating from different publicly available graph sets and reductions from other combinatorial hard problems. The benchmarks, which are publicly accessible through the PACE Challenge website \href{https://pacechallenge.org/2019/vc/vc_exact/}{Pace Benchmark}, consist of $200$ instances, each offering unique challenges and insights for the MCP. The set was collected from various sources: SAT benchmarks, BHOSLIB benchmarks, DIMACS benchmarks, and other sources\cite{pace-detail}.

Another intriguing benchmark that can be adapted for the MCP is the \href{https://github.com/t-dillon/tdoku}{SUDOKU Benchmark}. This benchmark comprises an extensive set of $ \sim 50000$ Sudoku puzzles. A Sudoku puzzle features a $9 \times 9$ grid, divided into nine $3 \times 3$ subgrids, with some cells pre-filled. The goal is to fill the remaining cells so that each row, column, and subgrid contains each digit from 1 to 9 exactly once. In the context of the MCP, the vertices represent specific numbers and squares of the Sudoku grid. An edge is formed between two vertices if the corresponding (number, square) pairs adhere to Sudoku's rules. Each Sudoku instance in this benchmark has a solution, offering a rich dataset for MCP research\cite{SANSEGUNDO20231008, aragon2018solving, prates2018problem}. Massive and sparse datasets are also used for testing algorithms, like GNN. Examples of such datasets are IMDB\cite{kriege2019unifying}, COLLAB\cite{yanardag2015deep}, TWITTER\cite{leskovec2014snap}, and FACEBOOK\cite{traud2012social}.

These benchmarks, with their inherent complexity and diversity, serve as excellent tools for researchers aiming to develop and refine algorithms for the MCP. Their varied nature allows for a comprehensive evaluation of algorithmic performance across different instance characteristics, thereby contributing significantly to the advancement of research in this domain.

\subsubsection{Quantum Benchmark}\label{qbenc}
In the Quantum Computing domain, the problem of benchmarks is one that goes beyond the specific case of MCP or MIS. Indeed the variety of experimental platforms, constraints, and ways of performing computation makes standardized benchmarking a painful (albeit often necessary) task. Regarding the specific problems we are analyzing in this review, we have seen how each platform works best for problems designed specifically to fit well its own connectivity but there are no reference benchmarks, albeit some theoretical results do exist\cite{liu2023computing,bojic2012quantum,cain2023quantum}.

In light of the diverse challenges associated with benchmarking in the quantum computing field—stemming from the variety of experimental platforms, operational constraints, and computational methodologies—we introduce a novel benchmark tailored to accommodate the capabilities of contemporary quantum computing hardware. This new benchmark is designed to bridge the gap in standardized benchmarking by providing a set of instances specifically crafted to challenge today's quantum computers. We release these instances, confident in their capacity to serve as robust benchmarks. The difficulty of these instances is primarily attributed to the minimal energy differences between the ground states (the planted solutions) and the first excited states. This subtle energy gap ensures that identifying the ground state becomes a sufficiently challenging task for quantum computing systems, thus testing their limits and highlighting their capabilities.

The intricacies of this hardness lie in the quantum computational principle of finding the lowest energy state, or the ground state, which corresponds to an optimal solution for the quantum algorithms. In classical computing, similar challenges are encountered in optimization problems where finding the global minimum in a landscape of local minima can be arduous. In the quantum context, the challenge is magnified by the great susceptibility of quantum systems to all sources of noise, making it difficult to distinguish between closely spaced energy levels and maintain the system in a highly coherent low-energy state. The instances we introduce are thus specifically designed to exploit these properties.

We extended these instances with small instances from the coding theory benchmark, the EVIL benchmark, and the DIMACS set.
The set of instances, all of them of order at most 128, and their corresponding details, including the sizes of the solutions, are made available at \href{https://github.com/zbogdan/qbenchmarks}{Quantum Benchmark graphs}, and a comprehensive list with the relative solution sizes is presented in Table \ref{qbenchmark}. By offering these instances to the quantum computing community, we aim to establish a benchmark that not only reflects the current state of quantum hardware but also pushes the envelope in quantum computing capabilities, driving the development of more sophisticated and powerful quantum systems. Importantly, as the size of quantum computers continues to increase, the problems in this benchmark can be quite easily scaled up as well to keep pace with the hardware technology.

\begin{table*}[h!]
\centering
\begin{minipage}{.5\linewidth}
\centering
\begin{tabular}{ |l|l|l|l| }
\hline
Name & $|\mathcal{V}|$ & $|\mathcal{E}|$ & $\omega(G)$ \\
\hline
g136x6 & 60 & 1716 & 48$^*$ \\
g136x8 & 80 & 3087 & 64$^*$ \\
g136x10 & 100 & 4827 & 60$^*$ \\
g136x12 & 120 & 6970 & 96$^*$ \\
g150x6 & 60 & 1735 & 54$^*$ \\
g150x8 & 80 & 3097 & 72$^*$ \\
g32559x6 & 66 & 2075 & 48$^*$ \\
g32559x8 & 88 & 3702 & 64$^*$ \\
g32559x10 & 110 & 5837 & 80$^*$ \\
g32565x6 & 66 & 2048 & 42$^*$ \\
g32565x8 & 88 & 3670 & 56$^*$ \\
g32565x10 & 110 & 5758 & 70$^*$ \\
g33896x4 & 72 & 2228 & 28$^*$ \\
g33896x5 & 90 & 3575 & 35$^*$ \\
g33896x6 & 108 & 5238 & 42$^*$ \\
g33896x7 & 126 & 7240 & 49$^*$ \\
g33918x4 & 72 & 2267 & 28$^*$ \\
g33918x5 & 90 & 3629 & 35$^*$ \\
g33918x6 & 108 & 5309 & 42$^*$ \\
g33918x7 & 126 & 7303 & 49$^*$ \\
g33920x4 & 76 & 2491 & 32$^*$ \\
g33920x5 & 95 & 4019 & 40$^*$ \\
g33920x6 & 114 & 5860 & 48$^*$ \\
g50634x3 & 60 & 1711 & 30$^*$ \\
g50634x4 & 80 & 3066 & 40$^*$ \\
g50634x5 & 100 & 4823 & 50$^*$ \\
g50634x6 & 120 & 6966 & 60$^*$ \\
1dc.64-c & 64 &  1473 & 10$^*$ \\
1dc.128-c & 128 &  6657 & 16$^*$ \\
\hline
\end{tabular}
\end{minipage}%
\begin{minipage}{.5\linewidth}
\centering
\begin{tabular}{ |l|l|l|l| }
\hline
Name & $|\mathcal{V}|$ & $|\mathcal{E}|$ & $\omega(G)$ \\
\hline
2dc.128-c & 128 &  2955 & 5$^*$ \\
1tc.64-c & 64 &  1824 & 20$^*$ \\
1tc.128-c & 128 &  7616 & 38$^*$ \\
1et.64-c & 64 & 1752 &18$^*$ \\
1et.128-c & 128 & 7456 &28$^*$ \\
1zc.128-c & 128 & 5888 & 18$^*$ \\
chv12x5 & 60 & 1525 & 10$^*$ \\
chv12x6 & 72 & 2247 & 12$^*$ \\
chv12x7 & 84 & 3124 & 14$^*$ \\
chv12x8 & 96 & 4147 & 16$^*$ \\
chv12x9 & 108 & 5281 & 18$^*$ \\
myc5x12 & 60 & 1674 & 24$^*$ \\
myc5x16 & 80 & 3032 & 32$^*$ \\
myc5x20 & 100 & 4747 & 40$^*$ \\
myc11x5 & 55 & 1288 & 10$^*$ \\
myc11x7 & 77 & 2624 & 14$^*$ \\
myc11x9 & 99 & 4442 & 18$^*$ \\
myc23x3 & 69 & 1759 & 6$^*$ \\
myc23x4 & 92 & 3403 & 8$^*$ \\
myc23x5 & 115 & 5539 & 10$^*$ \\
s3m25x2 & 50 & 910 & 8$^*$ \\
s3m25x3 & 75 & 2290 & 12$^*$ \\
s3m25x4 & 100 & 4274 & 16$^*$ \\
s3m25x5 & 125 & 6892 & 20$^*$ \\
C125.9 & 125 & 6963 & 34$^*$ \\
hamming6-2 & 64 & 1824 & 32$^*$ \\
hamming6-4 & 64 & 704 & 4$^*$ \\
johnson8-4-4 & 70 & 1855 & 14$^*$ \\
johnson16-2-4 & 120 & 5460 & 8$^*$ \\
\hline
\end{tabular}
\end{minipage}
\caption{Benchmarks for quantum algorithms  (optimal
values are marked with '$^*$').}
\label{qbenchmark}
\end{table*}

\section{Classical algorithms for MCP}\label{sec::classical_algo}

In the present section, we focus on results for the clique problem arising from 2014 that were based on previous research and those that extend the classical results. First, we detail the algorithmic landscape of the exact solvers and their background. Second, we will introduce shortly some recent results in the
heuristic approach.

\subsection{Exact algorithms}

From the historical point of view, classical algorithms, i.e., exact and heuristic algorithms, for solving the MCP have primarily focused on branch-and-bound approaches\cite{lawler1966branch}, or in other terms backtracking, with several significant advancements in recent years.  Previously, more general algorithms were known, such as the Bron–Kerbosch backtrack algorithm\cite{Bron1973}, which enumerated all maximal cliques and thus solved the MCP as a byproduct. Later, specific algorithms tailored for the maximum clique problem were developed. One of the early and influential algorithms for MCP was introduced by Carraghan and Pardalos\cite{carraghan1990exact}. This algorithm, known for its simplicity and effectiveness, employed a branch-and-bound method and was a good base to add pruning strategies in the following algorithms to reduce the search space. The algorithm performs a depth-first search from each vertex and prunes the search space by considering the number of remaining vertices that could potentially form a clique.

In the early years several basic ideas of auxiliary algorithms aiding the solution of the MCP were explored.  An early algorithm from Balas and Yu\cite{Balas1986} was mostly unnoticed due to its complexity, but contained many ideas that are very important nowadays -- for example usage of coloring the nodes of the graph. From recent years we can mention algorithm by Konc and Janežič\cite{Konc2007} that explores the usefulness of the concept of reordering the nodes, and another from Tomita\cite{tomita2017efficient} that take a more conservative approach. An especially interesting approach of the Russian doll technique was presented by \"Osterg\aa rd\cite{Ostergard2002}. Finally, would like to mention probably the first article about parallelization of the MCP problem\cite{Pardalos1998}.

A comprehensive overview covering classical methods prior to 2014 can be found in ref.\cite{wu2015review}. Here we shall further refer to the more recent papers.

\subsubsection{Auxiliary algorithms}

We shall describe the branch-and-bound algorithms later, here we would like to point out an important phenomenon concerning these. The running time of these algorithms is naturally closely connected to the search tree size. One factor is of utmost importance for the size of the search tree, and that is the branching factor\cite{golomb1965}. In the case of the state-of-the-art MCP solvers, this branching factor depends on the actual lower and upper bounds. The first is usually an incumbent solution, while the latter is obtained by an auxiliary algorithm. The most widely used is the coloring of the nodes of the graph. The bigger the gap between lower and upper bounds usually the bigger the branching factor and thus the size of the search tree. Thus, modern clique search algorithms pursue two goals: find a good incumbent solution fast and produce a tight upper bound during the
search.

A coloring of the nodes of the graph $G=(\mathcal{V},\mathcal{E})$ can be conveniently given by a map $f:\mathcal{V}\to\{1,\ldots,c\}$. Here the numbers $1,\ldots,c$ represent the colors and $f(v)$ is the color of the node $v\in \mathcal{V}$. The coloring of the nodes of a graph is called proper or legal coloring, and has two properties. First, each node receives exactly one color. Second, two nodes of the same edge cannot receive the same color. The minimum number of colors that a graph can be colored is called the chromatic number and is denoted by $\chi(G)$. Naturally, $\chi(G)\geq \omega(G)$ as for a clique of size $k$ one needs $k$ different colors for proper coloring. A greedy coloring is an upper bound to $\chi(G)$ and thus also an upper bound to the maximum clique size. The coloring may be used in the very beginning of the algorithm for the whole graph, or it can be performed at each node for a subgraph. Later takes more time, but has a stronger effect on the size of the search tree. The gain and balance between using coloring only on top or at each level, and also the complexity of the used coloring algorithm, including simplified partial coloring or recoloring using previous coloring results, is actively researched.

Such an upper bound proved so efficient that many state-of-the-art solvers use coloring or even stronger bounds on each level of search. Among others, these may include more complex coloring schemes as $b$-fold coloring, fractional coloring, and the Lovasz' theta function. For a detailed list and comparison of different upper bounds in clique search see Margenov et al.\cite{iet:/content/books/10.1049/pbpc024e_ch6}.

Another set of auxiliary algorithms, referred to as reduction rules, are used mainly for preprocessing purposes in order to decrease the size of the graph and/or complexity of the problem. These can be seen as polynomial time transformation of $G=(\mathcal{V},\mathcal{E})$ into $G'=(\mathcal{V}',\mathcal{E}')$ such that $G'$ has special properties. Mostly, these transformations fall into two categories:

\begin{enumerate}
\item $|V'|=|V|-1, \omega(G')=\omega(G)$
\item $|V'|\leq|V|, \omega(G')=\omega(G)-1$
\end{enumerate}

The first category is about a transformation where we can delete a node from the graph and preserve the clique number. An example of such a transformation is the dominance of two nodes. A node $u\in G$ can be deleted from the graph if there is a node $v$ present, such that $\partial u\subseteq \partial v$.

The second category is about a transformation where the size of the graph is reduced while the clique number is also reduced by one. An example of that is the so called vertex folding transformation. If a node $v$ has degree $N-3$, that is only two nodes $u,w$ are non neighbours to it, and they are adjacent, that is $\{u,w\}\in \mathcal{E}$, then one can remove $v,u,w$ and replace them by folding $u,v$ into $v'$ which has the common neighbourhood of $u,w$. (We would like to point out that vertex folding, among several other transformation methods, are special cases, and thus derivatives of the structure transformation\cite{EBENEGGER198483,ALEXE200327}.)

Similar transformations are widely used for fixed parameter tractable (FPT) algorithms and called kernelization in that field. The MCP does not admit an FPT algorithm, as it was detailed in the previous section, but the kernelization methods can be useful for problem reduction. The vast landscape of FPT algorithms and kernelization is out of the scope of this paper, we only refer to some previous works that are more connected to the underlying MCP\cite{BODLAENDER2009423,10.1145/3355502,10.1007/978-3-319-42634-1_28,Niedermeier2006-NIEITF,
  AKIBA2016211, Cygan2015}.

It is important to note that, branch-and-reduce algorithms, built upon kernelization methods, have proved to be extremely successful for the Vertex Cover problem. Moreover, the usual test graphs for the VCP are very different from the test graphs for the MCP. The hard instances for MCP usually have 500--1000 nodes and a density of 50-90\%, while the instances for VCP -- which should be considered as complement graphs to the instances for MCP -- have several thousands to even millions of nodes, with extremely small densities. 
Some researchers, such as those in ref.\cite{AKIBA2016211}, have indicated that these graphs differ significantly, leading to the conclusion that different algorithmic approaches are required for them. Specifically, kernelization is suitable for the typical VCP, whereas branch-and-bound is appropriate for the typical MCP. 
However, the PACE2019 competition demonstrated that these approaches could be successfully combined; after applying strong kernelization, the kernel of the problem should then be addressed using the branch-and-bound method. All three medalists employed this strategy\cite{dzulfikar2019pace}. Note, that problem setup of very spars graphs where the MCP is hard upon reason of the graph is being huge is a valid problem, and there are notable results in this area\cite{ChangLijun2019,ChangLijun2020,Rossi2015,doi:10.1137/1.9781611976472.10}. But in our paper we focus more on the small and hard instances.

Obviously, the presented two categories of transformations can be extended. In the first case, instead of deleting a node, one can aim to delete an edge, that is a transformation yielding to $|\mathcal{E}'|=|\mathcal{E}|-1$, as discussed in detail in Szabó et al.\cite{szabo2022b}. In the second case, one can propose increasing transformations, that is $|\mathcal{V}'|\geq |\mathcal{V}|$. Although they cannot be called as reductions, they can still result in either an easier problem or transform the graph into a bigger one that can be tackled by future reductions more efficiently, as described in Gellner et al.\cite{doi:10.1137/1.9781611976472.10}.
It is clear that this question requires more thorough research, and it is very possible that future state-of-the-art programs will include such reductions.

\subsubsection{Current achievements in Exact algorithms with
  branch-and-bound approaches}

Exact algorithms have played a pivotal role in advancing our ability to solve the maximum clique problem (MCP). Over the years, several methodologies have been proposed, refined, and optimized. From 2014 onwards, the field has seen significant advancements, particularly in the areas of branch-and-bound techniques.  Branch-and-bound methods have been at the core of most exact algorithms for solving the MCP. These algorithms explore the solution space by branching on vertices and bounding using stronger bounds, thereby systematically excluding some parts of the solution space that do not contain an optimal solution.

A branch-and-bound algorithm for the MCP typically starts by selecting a vertex, treating it as a single-vertex clique, and attempts to extend this clique by adding one vertex at a time from the remaining vertices. If the addition of a vertex results in a larger clique, the algorithm continues to explore this branch. However, if there are no more vertices to add, the algorithm backtracks and explores other branches. The bounding part of the algorithm comes from the ability to estimate an upper bound on the size of the maximum clique that can be obtained from a given branch. If this upper bound is smaller than the size of the largest clique found so far, this branch can be pruned. We refer to ref.\cite{wu2015review, carraghan1990exact, lawler1966branch} for an in-depth discussion on branch-and-bound approaches.

San Segundo and others in a series of papers from 2015 and 2016 enhanced their previous bit-parallel BB-MaxClique (BBMC) algorithm using several innovative techniques. While most of the ideas seem small, the overall result accumulates to a much better algorithm. They describe a technique of producing an upper bound that is actually better than the upper bound derived from the chromatic number. They call this bound Infra-Chromatic\cite{san2015infra}, which was improved in San Segundo et al.\cite{san2016improvedInfra} and enhanced in ref.\cite{san2016enhanced}. They combined this technique and thoroughly examined the effect with different vertex ordering strategies\cite{segundo2016improvedVertex}, and also introduced a very effective method of bit encoding for the coloring subprogram for efficient and fast chromatic bound search for the branches\cite{komosko2016fast}.

Based on previous results from 2010\cite{li2010efficient} Chu-Min Li, Hua Jiang, and others\cite{jiang2016combining} proposed for the MCP a  strategy named Large MaxClique (LMC) for the branch-and-bound methods has been established. They combined a preprocessing strategy to delete nodes that cannot belong to a maximum clique with the incremental MaxSAT reasoning for pruning branches. In ref.\cite{LI20171}, they extend these results by applying node ordering. The static vertex ordering in $G$ must be kept during the search and dynamic ordering of the vertices compared, and along with a version of mixed strategy three programs were evaluated (Static ordering MaxClique Solver -- SoMC, Dynamic ordering MaxClique Solver -- DoMC, Mixed ordering MaxClique Solver -- MoMC). In their next paper\cite{li2018incremental}, they investigate the result of searching for a large clique early, an incumbent solution, if that helps the algorithm (incremental upper bound MaxClique solver -- IncMC2). All proposed versions were superior in many cases to the other approaches at that time on random
and DIMACS instances as well.

An algorithm presented by Szabo and Zavalnij\cite{szabo2018different} uses several smaller contributions that add up to an algorithm, which is comparable to other state-of-the-art programs, sometimes slower, and for some instances still better than others up to date. The key points of the algorithm are the following. First, they use $k$-clique decision problem instead of MCP because that leads to better preconditioning and better branch factor. The MCP is solved by performing several $k$-clique search in an incremental manner. Second, they use a different reordering of the nodes in the branch opposite to other approaches, as the goal is not to find a better incumbent solution but to reduce the search tree, see also Ouyang\cite{Ouyang1998}. Third, they use a strong coloring scheme, incremental recoloring in the beginning, so to have the possibly best coloring for the graph. They also perform recoloring of nodes in each branch. Finally, they used an approach of constructing the branching node set described in detail in Szabó et al.\cite{SZABO2018118}.

Recently, in San Segundo et al.\cite{SANSEGUNDO20231008}, an exact algorithm, called CliSAT, to solve the MCP to proven optimality has been presented. This method is a combinatorial branch-and-bound algorithm, leveraging an advanced branching scheme enriched with two bounding strategies to minimize the branching tree size. The initial strategy employs graph coloring procedures and partial maximum satisfiability problems arising in the branching scheme. The second strategy incorporates a filtering stage utilizing constraint programming and domain propagation methods. CliSAT, tailored for structured and computationally challenging MCP instances, excels in dense environments characterized by numerous, interconnected large cliques. Rigorous testing on tough benchmark problems and newly identified challenging instances from diverse fields reveals CliSAT's superior performance over leading MCP algorithms, sometimes by several orders of magnitude. Specifically devised for demanding MCP problems with tens of thousands of vertices, these instances are marked by dense structures and multiple large, intertwined cliques. In this context, cutting-edge techniques are combinatorial branch-and-bound algorithms with bounding processes based on graph coloring and partial maximum satisfiability issues emerging from the branching strategy. CliSAT demonstrates remarkable results on the classical DIMACS benchmark, surpassing previous top-performing exact algorithms. Furthermore, in several categories of challenging instances, CliSAT achieves unparalleled performance, surpassing others significantly. CliSAT also shows competitive results for sparse graphs with as many as $150000$ vertices, despite not being primarily designed for such instances. 

CliSAT utilizes a constructive, $n$-ary branching approach, progressively constructing a clique by adding vertices one by one in a recursive manner. Within this context, $\hat{C} \subseteq \mathcal{V}(G)$ represents the clique linked to a specific branching node, comprising vertices included in $\hat{C}$ during previous branching steps. Additionally, each branching node corresponds with a subproblem graph $G'$, encompassing vertices that can be individually added to $\hat{C}$. Throughout its operation, CliSAT monitors the current best solution, referred to as $C_{inc}$. The lower bound, denoted as $lb$, is the size $|C_{inc}|$ of this solution. When a larger clique is identified during branching, signified by $|\hat{C}|>lb$, both $C_{inc}$ and $lb$ are updated. Post execution, $C_{inc}$ represents the maximal clique in $G$. The main idea of the branching scheme is to partition the vertex set of the subproblem graph $G'$ into two groups: i) the branching set $B$ and ii) the pruned set $P$ -- based on ref.\cite{iet:/content/books/10.1049/pbpc024e_ch6, SZABO2018118}.

Given the definition of $P$, at least one vertex from $B=\mathcal{V}(G')\setminus P$ is required to enhance the incumbent solution $C_{inc}$. Consequently, branching on vertices in $P$ at a certain node is unnecessary, leading the algorithm to backtrack when $B$ becomes empty. Once the pruned and branching sets are established, CliSAT executes a $|B|$-ary branching, generating a new branching node for each vertex in $B$ by including it in the current subproblem clique $\hat{C}$.
They also use strong incremental upper bounds for constructing the set $P$ as discussed in previous works like ref.\cite{li2018incremental,li2018new, san2015infra}.

Although scarce and considered extremely hard for combinatorial optimization, some examples of parallel algorithms were also published. Depolli et al.\cite{Depolli2013} used shared memory parallelization up to 24 cores, while Zavalnij\cite{Zavalnij2015} used distributed parallelization up to 512 cores. Finally, we would like to mention an interesting and unpublished concept by Hrga\cite{hrga2019} presented on the Zuze UG Workshop, which is apart from demonstrating a parallel solver for the MCP also using the Lovasz' Theta function as a bounding function. Note though, that the effective parallelization of the clique problem still remains mostly
open for future research.

Recent years have seen the development of precise algorithms for variations and extensions of the MCP. Papers\cite{jiang2017exact,jiang2018two,SANSEGUNDO201918} have detailed efficient algorithms for the maximum vertex-weighted clique problem. In this problem, one searches for a clique, where the sum of the weights on the nodes of the clique is maximal. The edge-weighted version of the problem has been addressed in works by San Segundo et al.\cite{san2019new} and Shimizu et al.\cite{shimizu2019branch}. A special case of the weighted problem, namely searching for maximum weight $k$-cliques in $k$-partite graphs arises from research on the protein docking problem\cite{ROZMAN2024}.  The authors in Furini et al.\cite{furini2021branch} have introduced exact algorithms for vertex and edge interdiction variants of the MCP. Lastly, an innovative exact algorithm for the knapsack problem with conflicts, which is akin to the MCP with an added knapsack constraint, was presented in Coniglio et al.\cite{coniglio2021new}. Recent research has also focused on identifying the clique number in large, sparse graphs typical of social networks. Algorithms tailored to these scale-free graphs, characterized by a power-law degree distribution, have been developed. These methods successfully solve the MCP with proven optimality in networks containing millions of vertices, as demonstrated in studies like Hespe et al.\cite{hespe2020wegotyoucovered} and Walteros et al.\cite{walteros2020maximum}. However, the techniques effective for these large, sparse graphs are generally not suitable for solving dense and challenging MCP instances.

To summarize the nuts and bolts of making an effective exact MCP solver we can highlight several key points. First, the effective combinatorial search was implemented as a simpler backtrack, and later improved by using branch-and-bound technique. Second, better upper bounds were introduced, such as proper coloring of the nodes. This direction of the research was later and still is continued nowadays whih more subchromatic bound such as Lovasz’ Theta function or MaxSAT reasoning. Third, the ordering of the nodes in the branch phase is researched, based on node degree, degeneracy or color index (number of neighboring color classes). Both approaches, such as finding a big clique fast that is using decreasing order in case of node degrees, and proving the maximum clique size that is using increasing order are actively researched. Fourth -- connected to the previously raised dichotomy --, the question of finding a good incumbent solution is raised. Fifth, the active use of preconditioning (kernelization) methods coming from the FPT research community as well as using such reductions during the search phase itself seems to play a major role in nowadays state-of-the-art algorithms.

\subsection{Heuristics for Hidden Cliques}

\subsubsection{Spectral and message passing algorithms}\label{sec::BP}

The discovery of hidden cliques in an Erd\H{o}s-R\'enyi graph $G$ of order $N$ and probability $p$ traces its origins to Ku{\v{c}}era's seminal research\cite{kuvcera1995expected}. Ku{\v{c}}era identified that, for sufficiently large constant $c$, a clique of size $c \sqrt{N \log N}$ will predominantly include the vertices with the most connections. This insight was further advanced by Alon et al.\cite{alon1998finding}, who applied spectral methods. Their study established that the optimal detection of a hidden clique in graph $G$ is feasible when the clique size $|C_{HC}|$is at least $c \sqrt{N}$, with $c$ being a sufficiently large constant.

The approach proposed by Alon (and colleagues) involves the following process. They start with the adjacency matrix $A$ of the graph $G$, sized $N \times N$, but with entries of $\pm 1$ rather than the typical $0,1$ (i.e., $0$ is replaced with $-1$, while $1$ remains unchanged). In scenarios where graph $G$ is generated according to the Erd\H{o}s-R\'enyi model with parameters $N$ and edge-formation probability $p=0.5$, it is improbable for matrix $A$ to have an eigenvalue exceeding $2\sqrt{N}$. However, if there is a clique $C_{HC}$ within $G$, the adjusted matrix $A$ will exhibit an eigenvalue of at least $|C_{HC}|-1$. Thus, if $|C_{HC}|$ is significantly greater than $2\sqrt{N}$, the largest eigenvalue of $A$ becomes a reliable indicator to differentiate between Erd\H{o}s-R\'enyi graphs and those containing hidden cliques. The eigenvector linked to this largest eigenvalue can then be utilized as an initial step in pinpointing the actual hidden clique $C$, although additional efforts are necessary to precisely locate it. Alon observed that the constant in the expression $|C_{HC}| = c\sqrt{N}$ can be optimized. By randomly selecting a fixed number $s$ of vertices from the graph $G$, there is approximately a $(|C|/|C_{HC}|)^s$ chance that these vertices form part of a clique $C$. Therefore, we can expect that after about $N^{s/2}$ tries, $s$ vertices from $C$ will be chosen. Focusing next on the subgraph $G'$, formed by the mutual neighbors of these vertices, it likely contains just $N/2^s$ vertices. This subgraph remains random and includes a nearly $c 2^{s/2} \sqrt{N/2^s}$ size clique. Consequently, in the context of $G'$, solving a hidden clique problem becomes more feasible with these improved parameters. This leads to the implication that for any $\epsilon > 0$, it is possible to identify planted cliques of size $\epsilon \sqrt{N}$ in a time complexity of $O(N^{\log 1/\epsilon})$\cite{feige2010finding}.

The leading approach for uncovering hidden cliques in random graphs (Erd\H{o}s-R\'enyi), backed by theoretical guarantees, is credited to the algorithm proposed by Deshpande et al.\cite{deshpande2015finding}. Their analysis, which becomes asymptotically exact as $N$ approaches infinity, demonstrates the algorithm's high probability of detecting hidden cliques of size $|C_{HC}| \ge \sqrt{N/e}$. This algorithm is an extension of belief propagation (BP), a heuristic approach in machine learning for estimating posterior probabilities in graphical models. BP has also been studied in statistical physics\cite{ li2021statistical, zhao2021belief, crotti2023matrix,  pan2021solving, caracciolo2021criticality, zhao2022equivalence, old2023generalized, mann2023belief, zhao2023local, xu2023generating, leisenberger2022fixing,nagano2023phase, aurell2023mighty}. As stated in ref.\cite{montanari2007solving,felzenszwalb2006efficient,yedidia2001generalized}, BP calculates marginal probabilities for individual nodes within a factor graph. It delivers precise results on tree structures and has also proven effective on graphs with loops, as indicated in ref.\cite{deshpande2015finding,mezard2009information,frey1998revolution,mooij2005sufficient}. As an iterative method, BP exchanges messages between links and nodes to ascertain marginal probabilities for each node. Once BP converges, indicating that marginal probabilities are determined, it sorts nodes based on these probabilities to solve the problem. Nonetheless, BP may fail or converge to a non-informative fixed point if the graph does not resemble a local tree. In BP's implementation for graphical models, each variable node represents a site in the original graph $G(\mathcal{V},\mathcal{E})$, while function nodes correspond to links in $G(\mathcal{V},\mathcal{E})$. For an in-depth understanding of the algorithm by Deshpande et al.\cite{deshpande2015finding}, readers are directed to their comprehensive paper. Here, we just recall the main steps.

The algorithm operates on a fully connected graph, characterized by an adjacency matrix $A$ with entries of $\pm 1$, similar to the spectral method introduced above. It functions through an iterative process, sending messages from links to nodes and then calculating specific values for each node. These values indicate whether a node is part of the hidden clique. This algorithm strikes a balance in terms of complexity and computational demands, positioning itself between simpler local methods, like greedy search approaches, and more comprehensive global strategies, such as the spectral techniques mentioned above. For our survey, we adapted a straightforward version of the algorithm from Deshpande et al.\cite{deshpande2015finding}, employing the equations they outlined. We briefly revisit these equations here:
\begin{equation}
  \label{messages}
  \begin{split}
   & \Gamma^{t+1}_{i \to j}=\log \frac{|C_{HC}|}{\sqrt{N}}+\sum^{N}_{l\neq i,j}\log\left( 1+\frac{(1+A_{l,i}) \text{e}^{\Gamma^{t}_{l\to i}}}{\sqrt{N}} \right) \\ &-\log\left( 1+\frac{\text{e}^{\Gamma^{t}_{l\to i}}}{\sqrt{N}}\right),
    \end{split}
  \end{equation}

\begin{equation}
  \label{marginal}
  \begin{split}
    &\Gamma^{t+1}_{i}=\log \frac{|C_{HC}|}{\sqrt{N}}+\sum^{N}_{l\neq i}\log\left( 1+\frac{(1+A_{l,i}) \text{e}^{\Gamma^{t}_{l\to i}}}{\sqrt{N}} \right)\\&-\log\left( 1+\frac{\text{e}^{\Gamma^{t}_{l\to i}}}{\sqrt{N}}\right),
     \end{split}
  \end{equation}
where $\Gamma^{t}_{i \to j},\Gamma^{t}_{j \to i},\Gamma^{t}_{i}  \in \mathbb{R}_{+}$.
  
Equations (\ref{messages}) and (\ref{marginal}) detail the dynamics of message exchanges and the computation of vertex parameters $\Gamma^{t}_{i}$ in a fully connected graph. This is facilitated by the adjacency matrix $A$, which accounts for both the presence and absence of links between nodes ($\pm1$ respectively). Logarithms are employed in these equations for enhanced numerical stability. The initial conditions for the messages in (\ref{messages}) are randomly assigned values less than zero.

The significance of $\sqrt{N}$ is noted as a critical scaling factor for problems involving hidden cliques.

Equation (\ref{messages}) is focused on updating the message sent from node $i$ to node $j$. This update is derived from all incoming messages to node $i$ from the previous iteration, excluding the message from node $j$ to node $i$. These messages are represented as odds ratios in real numbers $\mathbb{R}$, indicating the likelihood of node $i$ being part of the hidden set $C_{HC}$. The message from node $i$ to node $j$ conveys whether node $i$ is likely part of $C_{HC}$, based on the odds ratios from the other $N-2$ nodes in the graph, excluding nodes $i$ and $j$. A positive logarithmic difference in the sum suggests a higher probability of a node being in $C_{HC}$, while a negative difference implies a lower likelihood.

Equation (\ref{marginal}) deals with the numerical update of the vertex parameter $\Gamma^{t}_{i}$, calculated from all incoming messages to node $i$. It estimates the probability of node $i$ being in the hidden clique $C_{HC}$. Nodes more likely to be in the hidden clique will have a $\Gamma^{t_c}_{i\in C_{HC}}$ value greater than zero, while those unlikely to be in $C$ will have a $\Gamma^{t_c}_{i\not \in C_{HC}}$ value less than zero. $t_c$ is the time when the convergence is achieved.  As iterative BP  equations,  (\ref{messages}) and (\ref{marginal})  are useful only if they converge. The computational complexity of each iteration is $\mathcal{O}(N^2)$, indeed, equation (\ref{messages}) can be computed efficiently using the following observation:
 
 \begin{equation}
   \label{trick}
     \Gamma^{t+1}_{i \to j}=\Gamma^{t+1}_{i}-\log\left( 1+\frac{(1+A_{j,i}) \text{e}^{\Gamma^{t}_{j\to i}}}{\sqrt{N}} \right)+\log\left( 1+\frac{\text{e}^{\Gamma^{t}_{j\to i}}}{\sqrt{N}}\right).
   \end{equation}
 
The convergence of all messages and vertex quantities in the algorithm typically requires $\mathcal{O}(\log N)$ iterations. As a result, the overall computational complexity of the algorithm is $\mathcal{O}(N^2 \log N)$. When the messages in equation (\ref{messages}) reach convergence, the vertex quantities computed by equation (\ref{marginal}) are arranged in descending order. The top $h_{HC}$ elements are then selected for verification as a potential solution. If these elements constitute a solution, the algorithm concludes successfully; otherwise, it reports a failure. While this algorithm is theoretically exact as $N$ approaches infinity, its efficacy has been empirically validated on Erd\H{o}s-R\'enyi random graphs with planted solutions. In these tests, a hidden clique of size $C_{HC} \geq \sqrt{N/e}$ was embedded, and the numerical outcomes were consistent with the theoretical predictions.

\subsubsection{Monte Carlo Methods}

The application of Monte Carlo methods to the hidden clique problem represents an intriguing approach in computational graph theory. While Monte Carlo methods have been extensively used in various domains for probabilistic and numerical analysis, their application in detecting planted cliques within graphs is relatively recent. Jerrum\cite{jerrum1992large} highlighted a significant limitation of Monte Carlo methods (MCM) in this context, demonstrating that these methods cannot efficiently uncover hidden cliques of size $O(N^{0.5 - \epsilon}) $ with $ \epsilon > 0 $. This finding initially cast doubt on the efficacy of MCM, especially for cliques smaller than the square root of the graph's size. However, recent advancements have rekindled interest in this area. Both numerical and theoretical analyses have shown interesting results. From some observations in Montanari\cite{montanari2015finding}, Angelini\cite{chiara2018parallel} presented an accurate analysis on the performance of parallel tempering for the HCP. Parallel tempering\cite{hansmann1997parallel}, also known as replica exchange Monte Carlo, is a sophisticated computational technique used primarily in statistical mechanics and computational chemistry. The algorithm enhances the sampling efficiency of Monte Carlo simulations, especially in systems that experience slow dynamics due to energy barriers. The core concept of parallel tempering involves running multiple simulations (or replicas) concurrently at different temperatures. The idea of this method is to make configurations at high temperatures available to the simulations at low temperatures and vice versa. In her work, Angelini\cite{chiara2018parallel} applies parallel tempering to HCP, showing numerically that its time-scaling in the hard region is indeed polynomial for the analyzed sizes.

From a theoretical point of view, Gamarnik et al.\cite{gamarnik2019landscape} examined the landscape of sufficiently dense subgraphs of $G$ and their overlap with the planted clique. Employing the first-moment method, they provided evidence of a phase transition for the presence of the Overlap Gap Property (OGP)\cite{gamarnik2021overlap} when $|C_{HC}|$ is $\Theta(\sqrt{N})$. They presented the first concentration results for the $|C_{HC}|$-densest subgraph problem in the Erd\H{o}s-R\'enyi random graph $G(N,p=0.5)$, when $|C_{HC}|=N^{0.5-\epsilon}$, with $\epsilon>0$. Following this observation, Angelini et al.\cite{angelini2021mismatching} conducted an exhaustive numerical analysis demonstrating that a Monte Carlo algorithm of the same class used by Jerrum\cite{jerrum1992large} is generally suboptimal for recovering the planted clique. Specifically, they found that a Metropolis MC based on the posterior distribution is suboptimal in identifying a planted clique; such an algorithm does not succeed in polynomial time even when $|C_{HC}|$ is significantly larger than $O(\sqrt{N})$. However, they noted that a Parallel Tempering algorithm, when generalized by adding an inverse temperature parameter $\beta=1/T$ with values $\beta \in (0,\infty)$, can match the performance of Bayes optimal algorithms such as BP, effective in polynomial time down to $|C_{HC}|=\sqrt{N/e}$. More precisely, when the inverse of temperature $\beta \neq 1$, the Parallel Tempering algorithm achieves the performance level of BP, whereas with $\beta = 1$, the algorithm becomes suboptimal. This is attributed to the fact that MC algorithms work better at higher temperatures, where the global minimum of the free energy remains the planted solution, but entropic effects facilitate faster movement through the phase space, allowing the solution to be reached in less time.

These observations, namely those of Gamarnik et al.\cite{gamarnik2019landscape} and Angelini et al.\cite{angelini2021mismatching}, were corroborated in a recent publication by Chen et al.\cite{chen2023almost}. In this study, Chen et al.\cite{chen2023almost} revisited the original Metropolis algorithm proposed by Jerrum\cite{jerrum1992large}. They demonstrated that, contrary to other efficient algorithms which succeed when $\eta=0.5$ (as seen in $SM^1$-ES\cite{marino2023hard}), the Metropolis algorithm fails to recover a planted clique of size $|C_{HC}|=\Theta(N^{\eta})$ for any constant $0\leq\eta<1$. The authors reinterpreted Jerrum’s results, noting that Jerrum's findings indicate the existence of a starting state (instance-dependent) where the Metropolis algorithm does not find the planted clique within polynomial time. Furthermore, they argue that the Metropolis algorithm is ineffective over a broad range of temperatures when initiated from the most natural starting point—the empty clique. This addresses an open question posed in Jerrum's original paper\cite{jerrum1992large}. The paper also concludes that a simulated tempering variant of the Metropolis algorithm is equally ineffective in the same parameter regime. In contrast to these results, a recent pre-print by Gheissari et al.\cite{gheissari2023finding} demonstrates that a simple Markov Chain Monte Carlo (MCMC) approach, utilizing an energy function that relaxes that of Jerrum’s Metropolis process, can effectively find the planted clique down to the $|C_{HC}|=\sqrt{N}$ threshold in $O(N)$ steps, starting from a natural uninformed state. To achieve this, they relaxed the clique constraint for this process. This adaptation allows the Markov chain to evolve across all subsets of the full vertex set while incorporating a penalty term in the Hamiltonian for any missing edges. Moreover, they show that when initialized from the full vertex set both the gradient descent and its low-temperature MCMC analogue recover the planted clique in linearly many steps, as long as  $|C_{HC}|=\Omega(\sqrt{N})$. It's important to note that by relaxing the problem to the space of subgraphs, they introduce two intuitive uninformed starting points for the algorithm: the empty set or the full graph. Crucially, the selection of the starting point plays a significant role in the algorithm's success. Similar to the Metropolis process focused on cliques, they observe that starting with an empty set renders gradient descent ineffective in locating the planted clique.

\section{Learning algorithms for MCP: the graph neural networks (GNNs)}\label{sec::GNN_algo}

In recent decades, Artificial Intelligence (AI) has been a game-changer in various fields, thanks to its ability to process complex tasks and handle large datasets. Machine learning, a critical branch of AI, features algorithms that learn and make decisions or predictions from data. A more advanced subset of machine learning is Deep Learning (DL), renowned for its effectiveness in building and training multi-layered neural networks. DL has achieved notable success in areas dealing with extensive and intricate datasets\cite{lauri2023learning, marino2023solving, marino2021learning, chicchi2023complex}. One of its landmark achievements is AlphaFold\cite{jumper2021highly}, developed by DeepMind, which has made significant strides in solving the protein folding problem—a longstanding challenge in biology and in the realm of combinatorial optimization. Building upon recent advancements in DL, Graph Neural Networks (GNNs) have emerged as a groundbreaking tool in addressing combinatorial optimization challenges\cite{cappart2023combinatorial}. GNNs, uniquely structured to process graph-formatted data, are adept at making predictions related to nodes, edges, and various graph-centric tasks. Each GNN layer processes node embeddings and the adjacency matrix to produce updated node embeddings. This innovative architecture equips GNNs with the capability to efficiently tackle combinatorial issues, where the overall quality of a solution is often contingent on the collective configuration rather than isolated elements. Encouragingly, GNNs can be Turing universal in the limit\cite{loukas2019graph}, which motivates their use as general-purpose solvers.

The core idea behind GNNs is to learn how to incorporate the connections and features of both nodes and edges to make predictions or classifications. In these networks, each node gathers information from its neighbors through a process called message passing. During each iteration of this process, nodes aggregate and transform information from their neighbors, gradually updating their own state. The final node representations can then be used for various graph-level, node-level, or edge-level tasks. 

A GNN can be conceptually viewed as comprising layers, each represented by a function (activation function) $\mathbf{F}[\cdot]$ with parameters $\Phi$. This function processes node embeddings and the adjacency matrix to produce updated node embeddings. The network's operations can be expressed as:
\begin{equation}
\begin{split}
    &\mathbf{H}_1=\mathbf{F}[\mathbf{X}, A, \phi_0]\\
    &\mathbf{H}_2=\mathbf{F}[\mathbf{H}_1, A, \phi_1]\\
    &\vdots = \vdots\\
    &\mathbf{H}_K=\mathbf{F}[\mathbf{H}_{K-1}, A, \phi_{K-1}]\\
    & f[\mathbf{X}, A,\Phi]=\mathbf{F}[\mathbf{{H}_K \mathbf{1}}/N].
\end{split}
\end{equation}

Here, $\mathbf{X}$ denotes the input matrix (i.e., the vertex embedding matrix), $A$ is the adjacency matrix underlying the graph, $\mathbf{H}_k$ encompasses the evolved node embeddings at the $k^{th}$ layer, and $\phi_k$ (for $k=0,\dots,K-1$) represents the parameters of each respective layer. The function $f[\mathbf{X}, A, \Phi]$ signifies the network’s output, serving the learning objective, be it classification or regression. This output function is instrumental in defining the loss function for both supervised and unsupervised learning settings. For comprehensive insights on GNNs, references such as ref.\cite{zhou2020graph, prince2023understanding} offer detailed discussions.

In the realm of combinatorial optimization, GNNs demonstrate a remarkable ability to either forecast the qualities of various configurations or directly deduce solutions\cite{xu2019can, loukas2020hard, chen2020can, sato2020survey,garg2020generalization,barcelo2020logical, selsam2018learning, bengio2021machine, ozolins2022goal}. They undergo training using a series of problem instances with established optimal solutions. During this training phase, GNNs discern patterns and structures, enabling them to apply these insights to novel, unencountered instances. This learning mechanism positions GNNs as a powerful instrument in the field of optimization, offering a blend of precision and adaptability. For an introduction to the field of GNNs for combinatorial optimization we refer to Section 8.1.6 of ref.\cite{zhou2020graph}, here we review the main papers regarding MCP.

An unsupervised approach for MCP has been discussed in Karalias and Loukas\cite{karalias2020erdos}. In their paper, the authors introduce an unsupervised learning model that innovatively constructs a differentiable loss function. This function's minima are assured to yield low-cost, valid solutions to the problem at hand. Inspired by Erd\H{o}s's probabilistic method, this approach trains a GNN to generate a distribution over subsets of an input graph's nodes by minimizing a probabilistic penalty loss function. Crucially, this achievement is realized in a straightforward, mathematically grounded manner and such a loss acts as a certificate for the existence of a low-cost set\cite{karalias2020erdos}. It eschews the need for continuous relaxations, regularization, or heuristic adjustments to correct improper solutions. Successfully optimizing this loss function ensures the derivation of high-quality integral solutions that conform to the problem's constraints. Upon training completion, the authors apply a well-established technique from the realm of randomized algorithms. This technique is used to sequentially and deterministically extract a valid solution from the learned distribution. This method eliminates the need for polynomial-time reductions, directly addressing each problem instead. Experiments for MCP were conducted on the IMDB, COLLAB\cite{morris2020tudataset, yanardag2015deep} and TWITTER\cite{leskovec2014snap} datasets. Further experiments were done on graphs generated from the RB model\cite{xu2007random}, which has been specifically designed to generate challenging problem instances. The GNN shows high performance on such datasets.

Another innovative unsupervised approach is introduced in Min et al.\cite{min2022can}. This paper's objective is to utilize a geometric scattering-based GNN for efficiently approximating the solution to the MCP. The model stands out for its parameter efficiency, requiring only approximately $0.1\%$ of the number of parameters compared to earlier GNN baseline models. The method comprises three primary components. The first is a hybrid scattering-GNN model, which converts a set of basic node-level statistics into a probability vector $\vec{p}\in[0,1]^N$. This vector efficiently represents the likelihood of each node being part of the maximum clique. The graph scattering transform, central to this model, utilizes a wavelet-based approach for machine learning on graphs, as discussed in ref.\cite{gama2018diffusion, zou2020graph, gao2019geometric}. Different from the one-hop localized low-pass filters in certain GNNs that promote smoothness among neighboring nodes, these wavelets function as band-pass filters. They encapsulate long-range dependencies through the expansive spatial support of their aggregations. Conceptually, at each node, diffusion wavelets perform a comparative operation, assessing the difference between the averaged features of two differently sized neighborhoods. The second component is a constraint-preserving decoder, designed to identify the maximum clique. The ideal output for a graph $G$ with $N$ nodes is a representation $\vec{p}\in[0,1]^N$, where nodes within the maximum clique are assigned a value of $1$, while all others receive $0$. This allows for the extraction of the maximum clique using $O(N)$ operations. In practice, the model constructs a representation that assigns higher probabilities to nodes within the maximum clique and lower probabilities to others, facilitating the extraction of the maximum clique by selecting nodes with higher $p_i$ values. Experiments conducted on the IMDB, COLLAB, and TWITTER datasets demonstrate the high performance of the GNN, albeit with similar efficiency compared to the algorithm presented in Karalias et al.\cite{karalias2020erdos}.

The burgeoning debate in the field of artificial intelligence has recently raised the question of whether GNNs can rise to the challenge and supersede the performance of state-of-the-art classical algorithms such as greedy or message-passing strategies. An intriguing development in this discourse is the recent publication of the article\cite{schuetz2022combinatorial}. In this study, the authors explore the application of GNNs in addressing combinatorial optimization challenges. The core of their approach involves defining a Hamiltonian (cost function) $H$, representative of the optimization problem. This function is characterized by binary decision variables $x_i\in\{0,1\}$ ($i=1,\dots,|\mathcal{V}|$), each linked to a corresponding vertex $i$ within an undirected graph $G(\mathcal{V}, \mathcal{E})$. The graph structure effectively captures the interdependencies among these variables. To facilitate the use of GNNs, they employ a relaxation technique on the problem Hamiltonian, creating a differentiable loss function. This function is pivotal in training the GNN. Adhering to a recursive neighborhood aggregation framework, the GNN iteratively updates the feature vector of each node based on the aggregated information from its immediate neighbors. Upon completion of the unsupervised training phase, a projection heuristic is applied. This step is crucial for translating the continuous-valued outputs, denoted as $p_i$, back into binary decisions, $x_i\in\{0,1\}$. They demonstrate the efficacy of their methodology through numerical experiments focused on the Maximum Independent Set (MIS) and Max-Cut problems, showcasing promising results. In this study, the authors claim, \textit{"We find that the graph neural network optimizer performs on par or outperforms existing solvers, with the ability to scale beyond the state-of-the-art to problems with millions of variables"}\cite{schuetz2022combinatorial}. This assertion has sparked a flurry of responses in the AI community.

The initial critique of the work presented in Schuetz et al.\cite{schuetz2022combinatorial} was put forth by Stefan Boettcher\cite{boettcher2023inability}. Boettcher's analysis reveals that, upon detailed examination, the performance improvements of the GNN described in the original study are marginally better than those achieved through gradient descent. This GNN is outperformed by a greedy algorithm, particularly in the context of solving the Max-Cut problem. Similarly, the commentary by Angelini and Ricci-Tersenghi\cite{angelini2023modern} highlights another perspective. They demonstrate that a relatively simple greedy algorithm, which operates in near-linear time, can consistently produce higher-quality solutions for the MIS problem. Furthermore, this is achieved in a significantly shorter timeframe compared to the GNN approach discussed in Schuetz et al.\cite{schuetz2022combinatorial}. These critiques collectively suggest that while GNNs hold promise, their current efficacy, especially in comparison to more traditional algorithms like greedy methods, may be limited in certain combinatorial optimization scenarios. Responses from the authors to these comments can be found in references\cite{schuetz2023replym,schuetz2023reply}.

The final contribution to this ongoing discussion, including the comments previously mentioned, was provided by Gamarnik\cite{gamarnik2023barriers}. In this manuscript, Gamarnik introduces the OGP as a novel perspective for understanding the performance of GNNs. This work not only adds depth to the existing dialogue but also presents a groundbreaking approach in the analysis of GNNs within the context of combinatorial optimization problems. He has asserted that the presence of OGP significantly narrows the possibility for GNNs to surpass existing algorithms. In fact, he elucidates a fundamental constraint for deploying GNNs on random graphs, a constraint that holds true across a diverse spectrum of GNN architectures. This limitation pertains when the GNN's depth remains constant and does not scale proportionally with the size of the graph. This holds true irrespective of other facets of the GNN architecture, such as the internal dimension or update functions. However, it is worth mentioning that this does not exclude GNNs from being a valuable tool in the AI arsenal. While GNNs may face limitations in certain situations, their adaptability and learning potential position them as valuable tools for challenges where traditional algorithms falter. This is particularly true in dynamic environments or with complex, intricate graph data. Therefore, the decision to use classical algorithms or GNNs should focus on utilizing their respective advantages for varying problem types, rather than pursuing a universal solution. This situation highlights the richness of the methodologies available to navigate the complex realm of combinatorial optimization.

\section{Quantum Algorithms for MCP}\label{sec::quant_algo}

Quantum Computing\cite{nielsen2010quantum, albash2018adiabatic,chen2021review} has been proven to be theoretically capable of solving some computational problems in exponentially less time compared to the best-known classical algorithms. Although the exact computational complexity of quantum algorithms with respect to the classical ones is still an open problem, this has not stopped researchers from trying to beat classical algorithms using quantum resources. While it is impossible to be exhaustive about all the different approaches to quantum computing, here we will give a brief introduction to the three main techniques that are used when dealing with combinatorial optimization problems. Such techniques are: Quantum Minimum Finding algorithm (the circuit model)\cite{durr1996quantum}, Quantum Annealing\cite{kadowaki1998quantum} and Quantum Approximate Optimization Algorithm (QAOA)\cite{farhi2014quantum}. We will see later that all the results obtained in quantum computing regarding the MCP fall into one of these three categories.

Having established the theoretical groundwork and identified the primary quantum computing techniques for tackling combinatorial optimization problems, let's delve into the practical aspects of applying these methods. First of all, we assume to have access to a register of $N$ qubits, i.e. physical objects that exhibit an effective dynamics that can be described by the coherent evolution of a spin $1/2$ particle. Each one of the qubits has a state that can be represented in a standard quantum mechanical notation as $\ket{\psi} =  \alpha \ket{0} + \beta \ket{1}$ where $\ket{0}$ and $\ket{1}$ are the logical states of the system and $\alpha , \beta \in \mathbb{C}$ with $|\alpha|^2+|\beta|^2=1$. The $N$-qubit register can then be generally described by the state  $\ket{\psi}_N =  \sum_{i=0}^{2^N-1} \alpha_i \ket{i}$ where $\ket{i}$ is one of logical bitstrings that can be obtained using $N$ bits and again $\alpha_i \in \mathbb{C}$ with $\sum_i|\alpha_i|^2=1$. This system lives in a Hilbert space of dimension $2^N$ and can be manipulated by evolving its state with a carefully designed Hamiltonian $H$. Indeed, we know that the dynamics of a quantum system follow the Schrödinger equation $i\frac{d}{dt}\ket{\psi}_N=H\ket{\psi}_N$. It means that after a time $t$, the system will be in the state $\ket{\psi(t)}_N = e^{-itH}\ket{\psi(0)}_N$. Equivalently, we can say that the evolved system, under the application of a Unitary matrix $U$ ($U^\dagger U = \mathbb{I}$, where $^\dagger$ is the adjoint operator and $\mathbb{I}$ the identity matrix), will be $U=e^{-itH}$ and $\ket{\psi(t)}_N = U\ket{\psi(0)}_N$. In the following, we will then talk interchangeably of quantum computations performed by the action of a Hamiltonian $H$ or by the application of a Unitary matrix $U$ onto the state. 

Given this theoretical basis, without going into the details of the actual theory of Quantum Computation which is well beyond the scope of this review, it can be proved that with a careful choice of Unitaries $U$ (or Hamiltonians $H$) we can use the quantum representation $\ket{\psi}_N$ and leverage the power of having a large computational Hilbert space to solve some problems with algorithms faster than the best known classical counterparts. The most famous examples are: factoring of a number $N$ that, on a quantum computer, can be theoretically done in $\sim O(polylog(N))$\cite{shor1999polynomial} with respect to the classical  $\sim O(N)$; and searching of an element in an unstructured database of $N$ elements that takes  $\sim O(\sqrt{N})$ on a quantum computer\cite{grover1996fast}, while  $\sim O(N)$ on a classical one. 

From a practical standpoint, the endeavor to construct and precisely control a system of $N$ qubits, essential for realizing a Quantum Computer with exceptional capabilities, represents a significant ongoing challenge that is currently a focal point of both academic and industrial efforts. Indeed, there are prototypical quantum devices built from superconducting qubits, trapped ions, neutral atoms, photons, and quantum dots\cite{resch2019quantum} just to cite a few. Unfortunately, none of them has yet reached a stage where error-corrected algorithms can be run reliably and at a scale that achieves a significant advantage over the best classical algorithms. This is due to inherent noise\cite{harper2020efficient,martina2022learning} and coupling with the environment\cite{suter2016colloquium,buffoni2022third} that makes it difficult to scale them up. However, this does not imply that these devices, despite being imperfect and thus falling outside the theoretical guarantees of quantum information theory, cannot serve as effective heuristic solvers for certain problems. Therefore, it is hardly surprising that, despite the technical challenges involved in building a functional quantum computer, some theoretical research has shifted focus towards employing quantum techniques to address some of the most daunting NP-hard problems, including the MCP.

\subsection{The Circuit Model}
The circuit model of quantum computation is arguably the most popular and widespread model that realizes the capabilities of quantum information processing, described above. In this model of computation, the unitary $U$ that implements the computation is decomposed into combinations of single-qubit operations (or gates) and two-qubit entangling operations. It can be demonstrated that one can approximate with arbitrarily small error any $N$-qubit unitary by using only these two types of gates\cite{nielsen2010quantum}. In the circuit model of quantum computation, individual gates are sequentially applied to achieve the desired outcome, mirroring the way classical logic circuits are constructed from a combination of gates. This model is also often referred to as \textit{digital} quantum computation to distinguish it from other \textit{analog} approaches that we will also cover later.

The circuit model algorithm for the MCP is inspired by the Grover search algorithm\cite{grover1996fast}. Grover's algorithm can find an element by using a quantum circuit that, starting from the uniform superposition state $\ket{\psi}_N =  \frac{1}{\sqrt{2^N}}\sum_{i=0}^{2^N-1}\ket{i}$, creates constructive interference in the wavefunction to amplify the probability of measuring a marked state. This state is marked by a so-called Oracle, i.e., a reversible circuit that can verify if a certain state is a solution to our problem or not. The idea of using this Oracle-based model comes from the fact that solutions to NP-hard such as MCP and MIS are hard to find but can be verified in polynomial time. It would then be theoretically feasible to build a reversible circuit acting in polynomial time that verifies if a state is a solution of MCP, and marks it. One can then use the same logic as Grover's search to find the maximum clique of a graph with $|\mathcal{V}|$ vertices with a theoretical complexity of $\sim O(|\mathcal{V}|\sqrt{2^{|\mathcal{V}|}})$\cite{bojic2012quantum}. 

It has been demonstrated by subsequent works\cite{haverly2021implementation,bhaduri2023robust} that this algorithm could be implemented in devices using realistic constraints thus giving a promising outlook. So far, however, no real-world implementation of this algorithm has been put forward due to the limited coherence times of available quantum devices with respect to the required circuit length and the practical cost, in terms of the number of qubits and gates, that is required to actually build an Oracle\cite{haverly2021implementation}. 

Recently, theoretical works have tried to put forward an analog version of this algorithm based on Quantum Walks\cite{li2019algorithm,li2024parameter}. However, the degree to which these types of algorithms based on Quantum Walks are actually implementable on realistic hardware is still debated, and no experimental results are available.

\subsection{Quantum Annealing}
Quantum Annealing\cite{albash2018adiabatic,mcgeoch2022adiabatic} is one of the most common \textit{analog} forms of quantum computation. It is a computation where the desired final state $\ket{\psi(\tau)}$ is obtained by evolving a suitable initial $\ket{\psi(0)}$ by means of a proper time-dependent Hamiltonian $H(t)$. In Quantum Annealing, the problem to solve is encoded into a spin Hamiltonian:

\begin{equation}
    H_P = \sum_i \sigma^z_i + \sum_{i<j} J_{ij} \sigma^z_i \sigma^z_j,
    \label{eq:ising_ham}
\end{equation}

where $\sigma^z_i$ is the Pauli-Z operator acting of qubit $i$. Finding a solution to this problem means finding the ground state of $H_P$. To achieve this, we introduce a simpler Hamiltonian $H_I = \sum_i \sigma^x_i$ which possesses an easily preparable ground state.
This ground state serves as our initial state, $\ket{\psi(0)}$, from which we proceed to evolve the system using a time-dependent Hamiltonian:

\begin{equation}
    H(t) = \left (1- \frac{t}{\tau}\right ) \cdot H_I + \frac{t}{\tau} \cdot H_P,
    \label{eq:annealing}
\end{equation}

with $t\in[0,\tau]$. Calling $s=t/\tau$ the schedule parameter, \eqref{eq:annealing} becomes:
\begin{equation}
    H(s) = (1- s) \cdot H_I + s \cdot H_P.
\end{equation}
In order to end up with this protocol in the ground state of $H_P$, i.e., the correct solution to our problem, we can resort to the adiabatic theorem. Calling $E_0(s)$ and $E_1(s)$ the energies of the ground and first excited states of the Hamiltonian $H(s)$, the adiabatic theorem states that if the system is evolved in a time $1/\tau >> \min_{s\in[0,1]} (E_1(s) - E_0(s))^2$ it will remain in the local ground state of the Hamiltonian, meaning that measuring the resulting state $\ket{\psi(\tau)}$ will give the correct solution\cite{kadowaki1998quantum}.

In its general formulation, also called Adiabatic Quantum Computation, this way of performing a quantum computation can be proved to be equivalent to the circuit model. However, with the simple Hamiltonian presented above it is not and it can only be used as a heuristic solver for Quadratic Unconstrained Binary Optimization (QUBO) problems. Conveniently, MIS can be easily formulated in QUBO form as shown above. 

Indeed, given a graph $G = (V, E)$, the QUBO representation of MIS is
\begin{equation}\label{eq:QUBO_Hamiltonian}
    H = -A \sum_{i=1}^{N} x_i + B \sum_{(i,j) \in E} x_i x_j,
\end{equation}
At this point, it is sufficient to map the Hamiltonian (\ref{eq:QUBO_Hamiltonian}) onto the physical hardware of the quantum device and find the corresponding ground state using the proper algorithm.

In Chapuis et al.\cite{chapuis2017finding}, a D-Wave 2X Quantum Annealer has been tested for this purpose. Before discussing the results, it is convenient to divide the MIS instances into two classes. (i) problems that fit into the processor topology and (ii) problems that can't fit into the processor. 
Let's start from the first case, the D-Wave 2X processor has around 1000 qubits arranged in a peculiar connectivity which goes by the name of Chimera graph. This graph is quite sparsely connected, with a qubit being connected to 6 neighbors on average. This might seem like a huge limitation on the problems that we can solve using these types of devices, but there are workarounds. If the problem intended to be solved on the machine does not naturally conform to the Chimera graph structure—that is, it cannot be directly mapped onto the machine—one may resort to the technique of minor embedding as suggested by Vinci et al.\cite{vinci2015quantum}. In minor embedding, one uses multiple physical qubits to represent one logical qubit of the problem instance, thus expanding the available connectivity at a cost of some physical qubits. Once a minor embedding is found that effectively represents the desired graph on the Chimera graph, the qubits intended to embody the same logical qubit are linked together through a strong ferromagnetic interaction. This ensures they act collectively as a single, larger qubit. Consequently, these aggregated logical qubits are often referred to as \textit{chains}. Just to give a rough idea of how many resources one needs to do minor embedding, the largest fully connected graph that can be fit in the D-Wave 2X has $45$ nodes. 

Given this, the authors in Chapuis et al.\cite{chapuis2017finding} tested the performance of the D-Wave 2X on the MIS using random graphs with $45$ vertices and various edge probabilities $p \in [0.3, 0.9]$. The results of the quantum machine were then compared to various classical heuristic and exact algorithms for MIS like Simulated Annealing\cite{kirkpatrick1983optimization}, Fast Max-Clique Finder\cite{pattabiraman2015fast} and Gurobi optimizer. For these easy instances, all approaches found the correct MIS albeit in different runtimes, with the quantum approach having a mediocre performance and definitely no advantage. Scaling to larger graphs presents a wholly different challenge. To address this, researchers utilized $m$ edge contractions from Chimera graphs, forming what is referred to as the $\mathcal{C}_m$ family of graphs. This approach aimed to achieve precise embedding while preserving the largest possible graph that could be embedded into the machine. Performance comparisons were made with various heuristic solvers. For large graphs (containing more than 400 nodes), the D-Wave 2X frequently emerged as the sole system capable of identifying the Maximum Independent Set (MIS), accomplishing this significantly faster than Simulated Annealing. It achieved a speedup just shy of $10^6$ for graphs comprising $1000$ nodes. Although the improvement is specific to a particular family of graphs, achieving a six-order magnitude enhancement over classical solvers is undeniably impressive. This indicates the potential utility of heuristic quantum solvers for certain graph types. However, unlike the circuit model, there's no theoretical assurance of speedup, with only a few investigations, such as ref.\cite{kim2024quantum,bauza2024scaling}, exploring similar dynamics. Furthermore, it remains challenging to predict whether a specific instance will effectively benefit from quantum annealing. Indeed, a recent work\cite{cazals2025identifying}, carrying out annealing dynamics on a system composed of neutral atom qubits to solve MIS (albeit for much smaller instances of 80 qubits maximum), failed to find speedups for random instances of unit-disk graphs.

For problems that can't directly fit into a quantum processor, i.e., case (ii), the solution involves using classical branch-and-bound algorithms\cite{chapuis2017finding, pelofske2023comparing,pelofske2019solving}. These algorithms decompose the main problem into smaller, manageable subproblems, using classical branch and bound algorithms. Some of these subproblems are then processed on the quantum processor, and the results are combined to address the original problem comprehensively. This method is applied to challenges from the DIMACS benchmark, but not compared against other classical solvers. As such, while hybrid quantum-classical methods are viable and operational, it's still uncertain whether they offer advantages over entirely classical approaches. Moreover, in this case, quantifying the quantum solver's specific contribution compared to classical algorithms poses a challenge, rendering the quantum processor's utility in such hybrid setups ambiguous. Classical simulations of quantum annealing dynamics have shown potential in certain cases\cite{naghsh2019digitally}, suggesting an alternative avenue for exploration.

Despite some limitations, the application of Quantum Annealing for solving the MIS is promising, demonstrating the capability to tackle realistic instances on actual quantum devices with encouraging outcomes. Recent advancements in Quantum Annealing hardware have been remarkable, transitioning from devices with approximately $1000$ qubits and about $6$ couplings per qubit\cite{chapuis2017finding}, to the latest models boasting around $5000$ qubits, each with approximately $15$ couplings. This enhancement significantly increases the potential to embed fully connected graphs of up to 180 nodes. The deployment of this advanced architecture may offer promising prospects, as evidenced by some findings on related challenges\cite{povh2023advancing}. However, it remains to be determined whether these improvements confer a definitive advantage.

\subsection{Quantum Approximate Optimization Algorithm}

Another viable approach involves employing an optimized form of Quantum Annealing on a gate-model quantum computer, known as the Quantum Approximate Optimization Algorithm (QAOA)\cite{farhi2014quantum}. In QAOA, just like in Quantum Annealing, the solution to the problem is encoded in the ground state of a Hamiltonian.

In QAOA, the ground state is approximated through a variational approach, utilizing a specific ansatz and optimization process. We employ the same $H_I$ and $H_P$ Hamiltonian defined in Eqs.(\ref{eq:ising_ham}) and (\ref{eq:annealing}), respectively, for solving MIS  or MCP. The process begins from a uniform superposition state $\ket{\psi}_N =  \frac{1}{\sqrt{2^N}}\sum_{i=0}^{2^N-1}\ket{i}$, analogous to the initial state preparation in the circuit model approach. Then, we build our variational ansatz using two sets of free parameters $\gamma = \{\gamma_1, \dots, \gamma_p \}$ and $\beta = \{\beta_1, \dots, \beta_p \}$ as follows:

\begin{equation}
    \ket{\psi(\gamma,\beta)} = e^{-i\beta_p H_I}e^{-i\gamma_p H_P} \dots e^{-i\beta_1 H_I}e^{-i\gamma_1 H_P}\ket{\psi}_N.
    \label{eq:qaoa}
\end{equation}

After this quantum circuit's operation, the system undergoes measurement in the computational basis across multiple iterations. This process is essential for calculating the expectation value of the cost function:

\begin{equation}
    F_p(\gamma,\beta) = \bra{\psi(\gamma,\beta)}H_P\ket{\psi(\gamma,\beta)}\geq E_0,
\end{equation}

where $E_0$ is the ground state energy of $H_P$.
The parameters $\gamma$ and $\beta$ are optimized through a classical subroutine, aiming that, after an adequate number of iterations, the state $ \ket{ \psi(\gamma^*,\beta^*) }$ aligns with the ground state of $H_P$. This state represents the solution to our MIS problem.

For simplicity, one might consider QAOA as a digital counterpart to quantum annealing. Specifically, the state described in Eq.\eqref{eq:annealing} can be seen as equivalent to a trotterized version of quantum annealing dynamics, implemented with $p$ Trotter steps. To sum up this brief overview of QAOA, it's important to highlight that the algorithm functions as a hybrid between quantum and classical computing. The variational ansatz is executed on a quantum device, while the optimization of parameters is performed by a classical computer. This creates a synergistic loop between the classical and quantum domains to efficiently run the algorithm.

The implementation of QAOA for addressing MIS problems on a quantum processor utilizing Rydberg atoms (also known as neutral atoms) has shown promising results\cite{adams2019rydberg}. The groundbreaking research by Ebadi et al.\cite{ebadi2022quantum} demonstrated the successful resolution of MIS problems for a specific family of graphs, namely unit-disk graphs, with up to 80 vertices. These graphs are particularly well-matched to the connectivity of their quantum processor, akin to the favorable scenarios in quantum annealing. Comparative analysis with Simulated Annealing revealed a scaling advantage related to the hardness parameter of the instances, indicating a nearly quadratic speedup for the quantum approach. This represents a scenario where a theoretical quantum speedup can be substantiated, contrasting with the quantum annealing method where such proof is more elusive\cite{cain2023quantum}. 

Another recent study by Yin et al.\cite{yin2023solving} explores the use of Rydberg atom architectures for solving problems on a distinct class of graphs known as Kings graphs. Addressing the significant challenge of connectivity—a major hurdle for the practical deployment of quantum algorithms—an approach proposed by Nguyen et al.\cite{nguyen2023quantum} suggests a method for embedding problems of any connectivity with a quadratic overhead. Although this technique awaits experimental verification, it leverages the unique reconfigurability of Rydberg atom processors and their potential for qubit shuttling during computations\cite{vogel2019shuttling}. This adaptability positions Rydberg atom-based systems as an increasingly sought-after option in the quantum computing arena.

It is worth noting here that all algorithms for MCP implemented so far in the quantum realm are heuristics, hence, we have no general guarantees in terms of complexity or scaling. Regarding the implementation of exact classical or quantum algorithms in quantum processors, there's no speedup to be expected in this domain for quantum algorithms unless P=NP\cite{nielsen2010quantum,aaronson2010bqp}.

\section{Conclusions}\label{conclusion}

In this paper, we have embarked on a comprehensive exploration of the Maximum Clique Problem (MCP), delving into its formalism, theoretical bounds, and the plethora of algorithms—both classical, learning and quantum—developed to tackle it. Section~\ref{sec::MCP} laid the groundwork by defining MCP and outlining its complexity, setting the stage for an in-depth analysis of existing solution strategies. Following this, Section~\ref{sec::classical_algo} provided a thorough review of classical approaches to solving MCP, encompassing both exact algorithms and heuristic methods, and extending the discussion to Monte Carlo algorithms for the Hidden Clique Problem (HCP), aiming to cover the state-of-the-art in these domains.

In the realm of emerging technologies, Section~\ref{sec::GNN_algo} surveyed the advancements in Graph Neural Networks (GNNs) and their application to solving MCP, demonstrating the potential of leveraging machine learning techniques in combinatorial optimization.

The quantum computing frontier, covered in Section~\ref{sec::quant_algo}, presents the latest quantum algorithms for MCP, signifying a leap towards harnessing quantum mechanics for computational advantage in solving notoriously difficult problems. 

As quantum hardware evolves and becomes more capable of handling complex computations, the benchmarks outlined in this paper will serve as a critical resource for the community. They not only facilitate the rigorous testing of new algorithms but also encourage the development of more sophisticated and efficient solutions. By laying out the current landscape and future directions, this work aims to spur further research (both in quantum and classical) and collaboration in tackling MCP and similar combinatorial optimization problems.

As we have described above, the types of solvers covered in this review are very different and adapt well to different scenarios. Some of them are exact, some heuristic, some work well on large instances, some on small ones. This makes a cross-benchmark between all of these techniques meaningless for all practical purposes. In general, the choice of approach for any specific problem will depend on a variety of factors such as the scale of the problem, the structure of the graph (i.e. dense or sparse graphs), the available computational resources and the balance sought between provable accuracy and computational feasibility. We summarized the main takeaways for each algorithmic family in Table \ref{tab:comp}.

\begin{table*}[ht]
\centering
\label{tab:comp}
\caption{Comparison of different algorithmic approaches to the MCP.}
\renewcommand{\arraystretch}{1.2}
\begin{tabularx}{\textwidth}{|l|X|X|X|}
\hline
\textbf{Approach} & \textbf{Advantages} & \textbf{Limitations} & \textbf{When to Use} \\
\hline
\textbf{Exact Algorithms} &
Proven optimal solutions with strong performance on dense graphs (e.g., DIMACS benchmarks). &
Exponential time complexity, scaling is limited to graphs with thousands of vertices, even for advanced solvers. &
When optimality is required (e.g., verification, benchmarking) or for medium-sized dense graphs with up to $\sim 10^3$ vertices. \\
\hline
\textbf{Heuristic Algorithms} &
Fast, scalable to very large graphs, often sufficient for practical or real-time applications. &
May find only near-optimal cliques, cannot certify optimality, approximation hardness is provably strong. &
When approximate or ``good enough'' solutions are acceptable, especially in real-world large or sparse networks. \\
\hline
\textbf{GNN-Based Algorithms} &
Handling complex graph structures, can generalize patterns from similar instances, adaptable to dynamic or data-driven problems. &
Performance depends on training, limited theoretical guarantees, often outperformed by simple greedy algorithms on random graphs, depth limitations. &
For data-rich environments or earning-based combinatorial optimization, where many similar MCP instances are available. \\
\hline
\textbf{Quantum Algorithms} &
Theoretical potential for speedups if proper algorithms and hardware are combined. &
Current devices are noisy and small-scale, algorithms remain heuristic, no proven speedup. &
For experimental or small-scale instances and quantum benchmarking, especially to test new quantum hardware. \\
\hline
\end{tabularx}
\end{table*}

In conclusion, this review contributes to the foundational understanding of MCP and offers insights into the multifaceted approaches developed to address it. As the field of quantum computing continues to mature, it holds the promise of unlocking new possibilities in solving combinatorial optimization problems, potentially surpassing the limitations of classical computing. The journey from theoretical formalism to practical application underscores the dynamic and evolving nature of computational science, beckoning a future where these complex problems can be addressed more efficiently.

\section*{Data Availability}
The benchmark datasets presented in this paper and the code used to generate them are publicly available at \url{https://github.com/zbogdan/evil2} and \url{https://github.com/zbogdan/qbenchmarks}

\section*{Acknowledgments}

R.M. is supported by \#NEXTGENERATIONEU (NGEU) and funded by the Ministry of University and Research (MUR), National Recovery and Resilience Plan (NRRP), project MNESYS (PE0000006) "A Multiscale integrated approach to the study of the nervous system in health and disease" (DR. 1553 11.10.2022). L.B. was funded by PNRR MUR Project No. SOE0000098-ThermoQT financed by the European Union--Next Generation EU.

\section*{Author contributions}
RM reviewed the heuristics and graph neural network solvers. LB reviewed the quantum algorithms. BZ reviewed exact solvers and constructed the benchmarks. All authors contributed to the writing of the manuscript. 

\section*{Competing Interests}
The authors have no competing interests.

\bibliographystyle{unsrtnat}
\bibliography{references}

\end{document}